%% file: main.tex
\documentclass[10pt,twocolumn,letterpaper]{article}

\usepackage[pagenumbers]{cvpr} 
\usepackage{listings}
\usepackage{inconsolata}
\usepackage{microtype}
\usepackage[skins]{tcolorbox}
\usepackage{amsmath}
\usepackage{amsfonts}
\usepackage{adjustbox}
\usepackage{bm}
\usepackage{enumitem}
\usepackage{ulem}
\usepackage{soul}
\usepackage{bbding}
\usepackage{colortbl}
\usepackage{pifont}

\input{preamble}
\definecolor{cvprblue}{rgb}{0.21,0.49,0.74}
\usepackage[pagebackref,breaklinks,colorlinks,citecolor=cvprblue]{hyperref}

\newcommand{\customfootnotetext}[2]{{
  \renewcommand{\thefootnote}{#1}
  \footnotetext[0]{#2}}}

\title{Memory-Augmented Multimodal LLMs for Surgical VQA\\via Self-Contained Inquiry\vspace{-0.35cm}}

\author{
Wenjun Hou\textsuperscript{\rm 1,2}, 
Kaishuai Xu\textsuperscript{\rm 1$\ast$}, 
Yi Cheng\textsuperscript{\rm 1$\ast$}, 
Yan Hu\textsuperscript{\rm 2$\dagger$}, 
Wenjie Li\textsuperscript{\rm 1}, 
Jiang Liu\textsuperscript{\rm 2,3$\dagger$}\\
\textsuperscript{\rm 1}Department of Computing, The Hong Kong Polytechnic University \\
\textsuperscript{\rm 2}Research Institute of Trustworthy Autonomous Systems and Department of Computer Science \\and Engineering, Southern University of Science and Technology \\
\textsuperscript{\rm 3}School of Computer Science, University of Nottingham Ningbo China \\
{\tt\small houwenjun060@gmail.com, \{kaishuaii.xu, alyssa.cheng\}@connect.polyu.hk}\\
{\tt\small huy3@sustech.edu.cn, cswjli@comp.polyu.edu.hk, liuj@sustech.edu.cn}
\vspace{-0.5cm}
}
\begin{document}
\maketitle
\customfootnotetext{$\ast$}{Equal contribution. $\dagger$ Corresponding authors.}

\input{section/abstract}
\input{section/introduction}
\input{section/related_works}
\input{section/method}
\input{section/experiment}
\input{section/result}
\input{section/conclusion}

{
    \small
    \bibliographystyle{ieeenat_fullname}
    \bibliography{main}
}

\input{section/X_suppl}

\end{document}

%% file: preamble.tex
\usepackage{array}                
\usepackage{xspace}               
\usepackage{diagbox}              
\usepackage{multicol}             
\usepackage{multirow}             
\usepackage{tabularx}             
\usepackage{makecell}             
\usepackage{booktabs}             
\usepackage{arydshln}

\definecolor{pastelblue}{HTML}{A1C9F4}
\definecolor{pastelorange}{HTML}{FFB482}
\definecolor{pastelgreen}{HTML}{8DE5A1}
\definecolor{pastelred}{HTML}{FF9F9B}
\definecolor{pastelpurple}{HTML}{D0BBFF}
\definecolor{pastelgold}{HTML}{FFD700}
\definecolor{pastelpink}{HTML}{FF69B4}

\colorlet{pastelbluemuted}{pastelblue!50}
\colorlet{pastelorangemuted}{pastelorange!75}
\colorlet{pastelgreenmuted}{pastelgreen!75}
\colorlet{pastelredmuted}{pastelred!75}
\colorlet{pastelpurplemuted}{pastelpurple!75}
\colorlet{pastelgoldmuted}{pastelgold!75}
\colorlet{pastelpinkmuted}{pastelpink!75}

\DeclareRobustCommand{\hlomme}[1]{{\sethlcolor{pastelredmuted}\hl{#1}}}

\definecolor{red}{RGB}{255, 0, 0}
\definecolor{green}{RGB}{0, 176, 80}
\definecolor{purple}{RGB}{128, 0, 128}
\newcommand{\cmark}{\textcolor{green}{\ding{51}}} 
\newcommand{\xmark}{\textcolor{red}{\ding{55}}}
\newcommand{\smark}{\textcolor{purple}{\ding{77}}}

\def \margin {4pt}
\def \fmargin {-10pt}
\def \submargin {5pt}

%% file: section/abstract.tex
\begin{abstract}
Comprehensively understanding surgical scenes in Surgical Visual Question Answering (Surgical VQA) requires reasoning over multiple objects. Previous approaches address this task using cross-modal fusion strategies to enhance reasoning ability. However, these methods often struggle with limited scene understanding and question comprehension, and some rely on external resources (e.g., pre-extracted object features), which can introduce errors and generalize poorly across diverse surgical environments. To address these challenges, we propose \textsc{S$^2$Can}, a simple yet effective memory-augmented framework that leverages Multimodal LLMs to improve \underline{s}urgical context comprehension via \underline{S}elf-\underline{C}ont\underline{a}i\underline{n}ed Inquiry. \textsc{S$^2$Can} operates autonomously, generating two types of memory for context augmentation: Direct Memory (DM), which provides multiple candidates (or hints) to the final answer, and Indirect Memory (IM), which consists of self-contained question-hint pairs to capture broader scene context. DM directly assists in answering the question, while IM enhances understanding of the surgical scene beyond the immediate query. Reasoning over these object-aware memories enables the model to accurately interpret images and respond to questions. Extensive experiments on three publicly available Surgical VQA datasets demonstrate that \textsc{S$^2$Can} achieves state-of-the-art performance, offering improved accuracy and robustness across various surgical scenarios. \vspace{-0.4cm}
\end{abstract}

%% file: section/introduction.tex
\begin{figure}[t]
    \centering
    \setlength\belowcaptionskip{\fmargin}
    \includegraphics[width=0.88\linewidth]{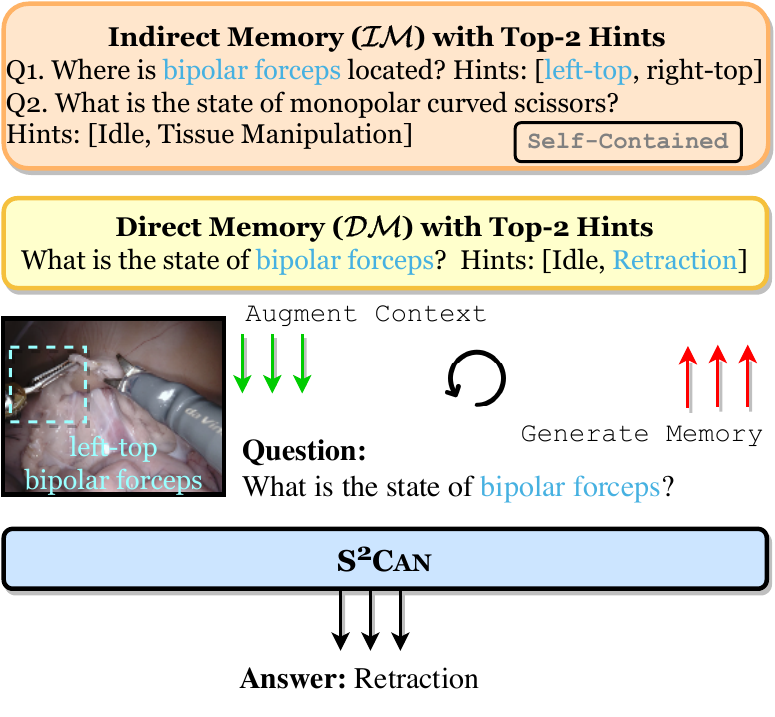}
    \caption{Given an image and a question, \textsc{S$^2$Can} generates memory via self-contained inquiry ({\color{red} $\rightarrow$}) and answers the question ({\color{green} $\rightarrow$}).}
    \label{figure: figure1}
\end{figure}
\section{Introduction}
Understanding surgical scenes is critical for analyzing egocentric surgical videos \cite{cholec80,cholect45,detection}, forming a foundational element in developing embodied agents for surgical contexts. This analysis encompasses various tasks, including action and phase recognition, tool detection, and tracking. Beyond these, visual question answering (VQA) \cite{vqa_iccv,san,mutan,block}, has emerged as a unified approach that can frame these tasks as a comprehensive problem of image- and video-grounded text generation. This paradigm not only streamlines the analytical process but also reduces the complexity of designing custom workflows for each task. Such integration is particularly advantageous in light of recent advancements in Large Language Models (LLMs), which can leverage vast resources to enhance both the efficiency and accuracy of surgical video analysis.

While medical VQA in pathology \cite{pathvqa} and radiology \cite{vqa-rad,slake} has garnered considerable attention, Surgical VQA \cite{sugvqa} that operates surgical videos remains less explored. Traditional medical VQA typically operates from a third-person perspective, focusing on answering questions based on static medical images. In contrast, Surgical VQA is an emerging field tailored specifically for egocentric surgical videos, addressing the unique challenges posed by the dynamic and complex nature of surgical environments. Previous research in VQA has primarily focused on fusion strategies \cite{sugvqa} for cross-modal representations, either by integrating object features \cite{ssgvqa,vqa_object1,vqa_object2} or by leveraging strong language priors \cite{suggpt}. However, these methods often struggle with scene understanding and question comprehension, relying on external resources such as scene graphs \cite{visual_genome} to improve accuracy. Unfortunately, these resources are not always accessible and may introduce errors when applied to diverse surgical contexts.

In this study, we explore the advanced capabilities of Multimodal LLMs (MLLMs) in understanding and analyzing complex surgical scenes. Comprehensively understanding a surgical image requires reasoning over multiple questions, as illustrated in Figure \ref{figure: figure1}. In this example, a set of object-related questions is first generated to serve as the model's memory, which can then be used to interpret any incoming questions about the image. This memory is subsequently refined to align with user-provided questions. When asked "What is the state of the bipolar forceps?", reasoning over these object-aware memories enables the model to comprehend both the position of the bipolar forceps and the state of other tools, thus enhancing its ability to understand the scene and provide an accurate answer. In addition, despite the strong reasoning capabilities of MLLMs, directly answering surgical questions remains challenging. While MLLMs may recall the correct answers, they sometimes rank them incorrectly. Presenting relevant candidate answers could help the model better interpret both the image and the question, thereby improving its overall accuracy.

To better utilize the contextual information of an image and address the challenges outlined, we introduce a simple yet effective memory-augmented MLLM designed to enhance \underline{s}urgical scene understanding via \underline{S}elf-\underline{C}ont\underline{a}i\underline{n} Inquiry, which we refer to as \textsc{S$^2$Can}. Our approach leverages the inherent strengths of MLLMs in a self-sufficient manner by generating memory entries that capture essential contextual information derived from surgical images. Unlike other multimodal learning methods that require external resources, our approach generates memories through a self-contained inquiry mechanism, effectively eliciting the full capabilities of MLLMs. Specifically, \textsc{S$^2$Can} generates two types of memory: \textit{Direct Memory} (DM), produced by generating hints related to the user-provided question, and \textit{Indirect Memory} (IM), created by prompting MLLMs to generate relevant questions with hints. DM provides immediate information to answer the question, while IM contains valuable context about the surgical scene, independent of the given question. These memory entries are subsequently utilized as contextual information in responding to new questions, thereby enabling the model to deliver more precise answers. To summarize, our contributions can be summarized as three-fold:
\begin{itemize}
    \item We propose \textsc{S$^2$Can}, a simple yet effective memory-augmented MLLM, designed to enhance the contextual understanding of surgical scenes and related questions.
    \item We introduce a self-contained inquiry mechanism that prompts MLLMs to generate question-hint pairs relevant to given surgical images, serving as contextual information for VQA. This mechanism fully leverages the MLLM's reasoning capabilities without requiring external data or resources.
    \item We conduct extensive experiments and provide a comprehensive analysis of \textsc{S$^2$Can}, alongside multiple MLLMs, on three publicly available Surgical VQA datasets. Our results demonstrate that \textsc{S$^2$Can} achieves state-of-the-art (SoTA) performance, advancing the field of surgical VQA. 
\end{itemize}

%% file: section/related_works.tex
\section{Related Works}
\subsection{Medical and Surgical VQA}
Medical VQA \cite{aioz_mevf_miccai19} has been extensively studied across various modalities, including radiographs \cite{slake,vqa-rad,ehrxqa}, pathology images \cite{pathvqa}, and others \cite{ImageCLEFVQA-Med2019,pmc_vqa}. Notably, \citet{cond_reason} proposed conditional reasoning for answering medical queries, while \citet{m3ae} introduced a multi-modal masked autoencoder pretrained on large-scale data, achieving strong results across multiple VQA datasets. \citet{pmc_vqa} further advanced medical VQA via visual instruction tuning, showing significant improvements. Other key contributions in the field include \cite{MUMC,pubmedclip,mtl}. In contrast, surgical VQA remains underexplored. One representative research \cite{sugvqa} in this field is adapting the VisualBERT \cite{visualbert} to surgical scenes with a residual MLP module. They also released three VQA datasets, i.e., EndoVis-18-VQA, Cholec80-VQA, and PSI-AVA-VQA. In detail, EndoVis-18-VQA is annotated from the MICCAI Endoscopic Vision Challenge 2018 dataset \cite{allan20202018roboticscenesegmentation}, Cholec80-VQA is annotated from the Cholec80 dataset \cite{cholec80}, and PSI-AVA-VQA dataset is extracted from the PSI-AVA dataset \cite{psi-ava}. More details can be found in Table \ref{table: data_stat}. Subsequently, \citet{suggpt} leveraged GPT-2 \cite{gpt2} to enhance performance, while \cite{surgicalvqal,surgicalvqal++} focused on target localization, annotating two VQAL datasets, i.e., EndoVis-18-VQAL and EndoVis-17-VQAL. \citet{ssgvqa} further advanced surgical VQA by incorporating scene graphs for question answering.

\begin{figure*}[t]
    \centering
    \setlength\belowcaptionskip{\fmargin}
    \includegraphics[width=0.9\linewidth]{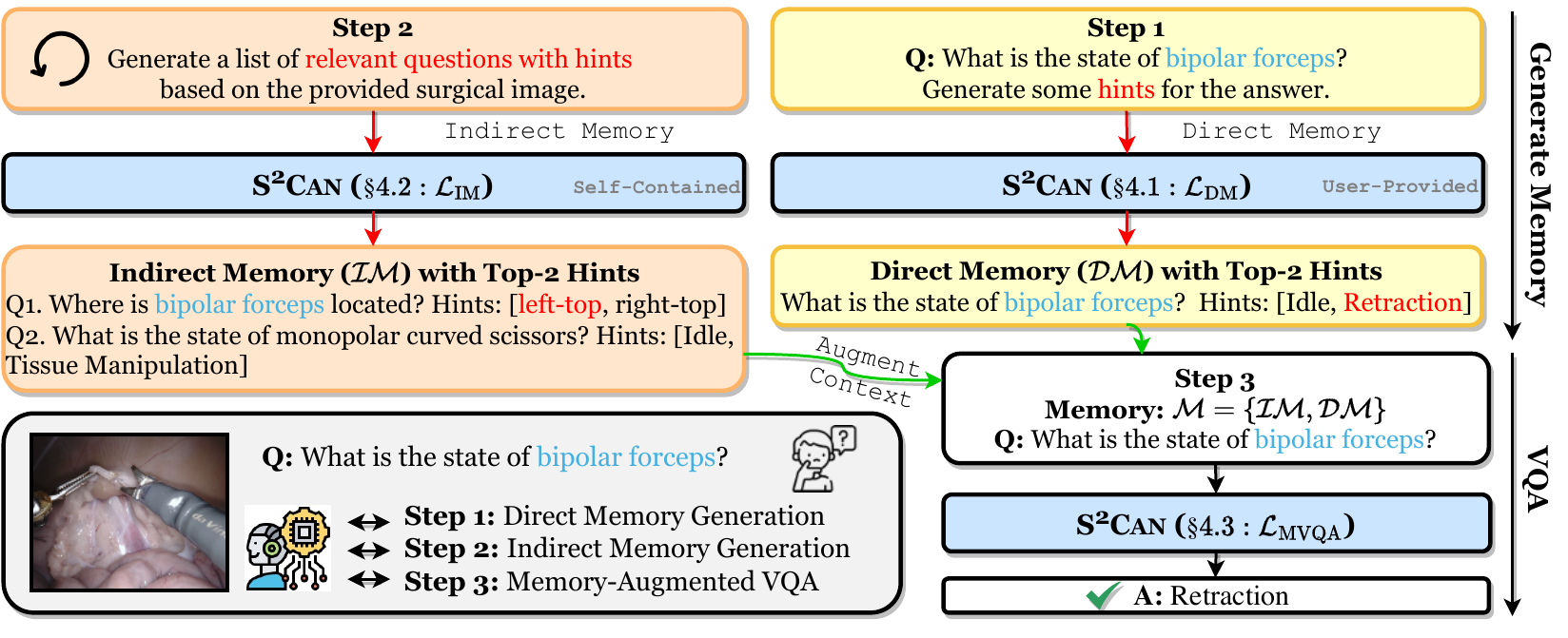}
    \caption{Illustration of our proposed \textsc{S$^2$Can} framework, which first generates memory and then utilizes it for VQA. {\color{red} Red spans} provide highly-related information for the answers. Note that Indirect Memory can be used to enhance any relevant questions of the given image.}
    \label{figure2}
\end{figure*}

\subsection{Multimodal LLMs and their Applications}
Various MLLMs have been developed using different LLMs with diverse capabilities \cite{2023GPT4VisionSC,minigpt}. \citet{blip2} proposed bootstrapping LLMs to enhance their multimodal abilities, while \citet{llava} introduced visual instruction tuning to improve LLMs' understanding of visual inputs. \citet{qwenvl} developed Qwen-VL, a model with strong capabilities in both understanding and generation and \citet{blip3} released a family of multimodal models demonstrating promising performance. Regarding domain-adapted MLLMs, medical models have shown significant potential across various modalities, such as radiology and pathology. \citet{llavamed} introduced a medical version of LLaVA \cite{llava}, while \citet{medflamingo} developed Med-Flamingo, based on Flamingo \cite{flamingo}, incorporating multiple cross-attention layers. \citet{huatuogptvision} further advanced this field by collecting larger alignment and tuning datasets, resulting in HuatuoGPT-Vision. Various applications \cite{r2llm,rule} have been built on top of these MLLMs. However, the use of MLLMs to improve surgical VQA remains rare, with \citet{llava_surg} developing an LLaVA model specifically for surgical video analysis.

%% file: section/method.tex
\section{Preliminaries}
\subsection{Problem Formulation}\label{problem_form}
Given a surgical image $\mathcal{I}$ and a question $\mathcal{Q}$, optionally accompanied by context $\mathcal{C}$, a MLLM parameterized by $\theta$ generates an answer $\mathcal{A}$ based on the image and question. The probability of the answer $\mathcal{A} = \{a_1, \dots, a_T\}$ and its corresponding loss function $\mathcal{L}_\mathcal{A}$ are defined as follows:
\begin{equation}
\setlength{\belowdisplayskip}{\margin}
\setlength{\abovedisplayskip}{\margin}
    \begin{split}
    p(a_t) &= \text{MLLM}_{\theta}(\mathcal{I}, [\mathcal{Q}; \mathcal{C}], \mathcal{A}_{<t}), \\
    p(\mathcal{A}) &= \prod_{t=1}^{T} p(a_t), \\
    \mathcal{L}_\mathcal{A} &= -\log p(\mathcal{A}) = -\sum_{t=1}^{T} \log p(a_t),
    \end{split}
\end{equation}
where $a_t$ is the $t$-th token in the answer, and $\mathcal{C}$ can be pre-extracted objects, clinical attributes, or memory that will be detailed later.

\subsection{The \textbf{\textsc{S$^2$Can}} Framework}
A typical autoregressive MLLM comprises three main components: a visual encoder ($\mathtt{Encoder}$), a visual connector ($\mathtt{Resampler}$), and an $\mathtt{LLM}$. Specifically, the visual encoder processes an input image, converting it into visual representations. These representations are then mapped by the visual connector to the embedding space of the LLM. Subsequently, the LLM generates a text sequence based on these embeddings. Specifically, given an image $\mathcal{I}$ and an input $\mathcal{X}$, the process of prompting \textsc{S$^2$Can} can be formulated as follows:
\begin{equation}
\setlength{\belowdisplayskip}{\margin}
\setlength{\abovedisplayskip}{\margin}
    \begin{split}
        \textsc{S$^2$Can}(\mathcal{I}, \mathcal{X})&=\left\{
        \begin{aligned}
            \bm{h}_{\mathcal{I}}&=\mathtt{Encoder}(\mathcal{I}), \\
            \bm{h}_{\mathcal{L}}&=\mathtt{Resampler}(\bm{h}_{\mathcal{I}}), \\
            \mathcal{Y}&=\mathtt{LLM}(\bm{h}_{\mathcal{L}}, \mathcal{X}), \\
        \end{aligned}
        \right. \\
    \end{split}
\end{equation}
where $\bm{h}_{\mathcal{I}}$ represents the hidden visual representations produced by the $\mathtt{Encoder}$, $\bm{h}_{\mathcal{L}}$ denotes the embeddings generated by the $\mathtt{Resampler}$, and $\mathcal{Y}$ is the output produced by \textsc{S$^2$Can} in response to the input question.

\section{Methodology}
In this section, we outline three steps of \textsc{S$^2$Can}, as shown in Figure \ref{figure2}: (1) Direct Memory Generation via Hint Generation, (2) Indirect Memory Generation via Self-Contained Inquiry, and (3) Memory-Augmented Surgical VQA.

\subsection{Direct Memory Generation}\label{direct_memory}
\textbf{Hint Generation.} A surgical-related question, particularly one asking about objects or their attributes, typically corresponds to a single correct answer. Despite the strong capabilities of MLLMs, generating an accurate answer directly can be challenging, especially for complex questions and scenes. To address this issue, we introduce a hint-generation method to augment contextual information based on a user-provided question, as shown in the top-right part of Figure \ref{figure2}. Instead of providing a single, definite answer to $\mathcal{Q}$, which could potentially be erroneous, we leverage \textsc{S$^2$Can} to generate a list of hints $\mathcal{H} = \{\mathcal{H}_1, \dots, \mathcal{H}_K\}$. By incorporating these hints into the VQA process, we bootstrap the model, allowing it to improve its ability to rank these candidates more accurately. Specifically, given $\mathcal{I}$, $\mathcal{Q}$, and a prompt $\mathcal{P}_\mathcal{DM}$ for hint generation, this process is formulated as:
\begin{equation}
\setlength{\belowdisplayskip}{\margin}
\setlength{\abovedisplayskip}{\margin}
\mathcal{H} = \{\mathcal{H}_1, \dots, \mathcal{H}_K\} = \textsc{S$^2$Can}(\mathcal{I}, [\mathcal{Q}; \mathcal{P}_\mathcal{DM}]).
\end{equation}
We concatenate the question and $K$ hints to form the \textit{Direct Memory}, represented as $\mathcal{DM} = \{(\mathcal{Q}, \mathcal{H})\}$.

\noindent\textbf{Hint Annotation.} To annotate hints for each question, we first collect all answer candidates from the training data. Next, we select the top $K-1$ candidates based on their frequency. Finally, the ground-truth answer is appended to this list of candidates to form the complete hint set.

\subsection{Indirect Memory Generation}\label{indirect_memory}
\textbf{Self-Contained Inquiry.} A surgical image may correspond to multiple question-answer (QA) pairs that, while related, address different aspects of the scene. Reasoning over these QA pairs helps the model better interpret the image. Since this contextual information can be inferred by MLLMs from the image alone, without the need for external data, we refer to these QA pairs as \textit{Indirect Memory} ($\mathcal{IM}$). To prevent the model from memorizing answers during training, we convert the answers into hints using the same method described in \S \ref{direct_memory}. As a result, $\mathcal{IM}$ becomes a list of question-hint pairs. Specifically, given an image $\mathcal{I}$ and a self-contained inquiry $\mathcal{P}_\mathcal{IM}$, the model generates:
\begin{equation}
\setlength{\belowdisplayskip}{\margin}
\setlength{\abovedisplayskip}{\margin}
\mathcal{IM} = \textsc{S$^2$Can}(\mathcal{I}, \mathcal{P}_\mathcal{IM}),
\end{equation}
where $\mathcal{IM} = \{(\mathcal{Q}^{\mathcal{I}}_1, \mathcal{H}^{\mathcal{I}}_1), \dots, (\mathcal{Q}^{\mathcal{I}}_N, \mathcal{H}^{\mathcal{I}}_N)\}$ represents the generated list of question-hint pairs. These pairs form the indirect memory, enriching the model's contextual understanding of the scene.

\noindent\textbf{Indirect Memory Selection.} We select the top $M$ entries from $\mathcal{IM}$ for each question based on cosine similarity:
\begin{equation}
\setlength{\belowdisplayskip}{\margin}
\setlength{\abovedisplayskip}{\margin}
    \mathcal{IM}_\mathcal{Q} = \operatorname{Top}_M \cos{(f_\mathcal{Q}, f_{\mathcal{Q}_n})},
\end{equation}
where $f_*$ represents the TF-IDF \cite{tfidf} features, and $\mathcal{IM}_\mathcal{Q}$ denotes the selected indirect memory for question $\mathcal{Q}$. During training and inference, the given question is excluded from the indirect memory to avoid information leakage.

\noindent\textbf{Indirect Memory Annotation.} Indirect memory is not directly relevant to a given question but instead relevant to an image. Thus, we take the annotated QA pairs as indirect memory annotations. To be specific, for an image with multiple QA pairs, we select samples with $\#\mathcal{Q} \geq N$ in the training data. Once we collect these annotations for each image, we further sort them by the frequency of questions and finally, these QA pairs are concatenated into a sequence for training to form the indirect memory.

\subsection{Memory-Augmented Surgical VQA}\label{mvqa}
The model’s memory is updated by integrating both indirect and direct memories, resulting in the composite memory set $\mathcal{M} = \{\mathcal{IM},\mathcal{DM}_\mathcal{Q}\}$. Reasoning over this combined memory structure enhances the model's contextual understanding and improves its ability to generate accurate and relevant outputs. Once the memory is created, we prompt \textsc{S$^2$Can} with the image $\mathcal{I}$, the question $\mathcal{Q}$, and the memory $\mathcal{M}$ to produce an answer $\mathcal{A}$, which can be modeled as:
\begin{equation}
\setlength{\belowdisplayskip}{\margin}
\setlength{\abovedisplayskip}{\margin}
\mathcal{A}=\textsc{S$^2$Can}({\mathcal{I}}, [\mathcal{M}; \mathcal{Q}]).
\end{equation}

\input{section/figures_tables/exp_results}

\subsection{Training}
Our \textsc{S$^2$Can} is trained to generate direct memory, indirect memory, and the answer to the given images and questions. Although these three tasks have different objectives, they can all be framed as autoregressive generation tasks and optimized using a negative log-likelihood loss, as described in \S\ref{problem_form}. Specifically, given a batch of training samples, the overall loss function $\mathcal{L}$ is formulated as:
\begin{equation}
\setlength{\belowdisplayskip}{\margin}
\setlength{\abovedisplayskip}{\margin}
    \mathcal{L} = \mathcal{L}_\text{DM} + \mathcal{L}_\text{IM} + \mathcal{L}_\text{MVQA},
\end{equation}
where $\mathcal{L}_\text{DM}$ (\S\ref{direct_memory}), $\mathcal{L}_\text{IM}$ (\S\ref{indirect_memory}), and $\mathcal{L}_\text{MVQA}$ (\S\ref{mvqa}) represent the direct memory generation loss, indirect memory generation loss, and memory-augmented VQA loss, respectively.

\subsection{Inference}
After training, we can leverage the multi-round prompting capability of \textsc{S$^2$Can} as described earlier. Given an image $\mathcal{I}$ and a user-provided question $\mathcal{Q}$, the inference process can be formulated as follows:

\begin{equation}
\setlength{\belowdisplayskip}{\margin}
\setlength{\abovedisplayskip}{\margin}
    \begin{split}
        \text{Inference}(\mathcal{I}, \mathcal{Q}) &= \left\{
        \begin{aligned}
            \mathcal{DM} &= \textsc{S$^2$Can}(\mathcal{I}, [\mathcal{Q}; \mathcal{P}_\mathcal{DM}]), \\
            \mathcal{IM}_\mathcal{Q} &= \textsc{S$^2$Can}(\mathcal{I}, \mathcal{P}_\mathcal{IM}), \\
            \mathcal{M} &= \{\mathcal{DM}, \mathcal{IM}_\mathcal{Q}\}, \\
            \mathcal{A} &= \textsc{S$^2$Can}(\mathcal{I}, [\mathcal{M}; \mathcal{Q}]). \\
        \end{aligned}
        \right.
    \end{split}
\end{equation}
The inference pipeline, denoted as $\mathcal{A} = \mathtt{Inference}(\mathcal{I}, \mathcal{Q})$, first generates both direct memory ($\mathcal{DM}$) and indirect memory ($\mathcal{IM}_\mathcal{Q}$) and then reasons over the image and the memory and answers the question.

%% file: section/figures_tables/exp_results.tex
\begin{table*}[t]
    \centering
    \resizebox{\textwidth}{!}{
    \begin{tabular}{l|c|ccc|ccc|ccc}
    \hline
    \multirow{2}{*}{\textbf{Model}} & \multirow{2}{*}{\textbf{LM}} & \multicolumn{3}{c|}{\textbf{EndoVis-18-VQA}} & \multicolumn{3}{c|}{\textbf{EndoVis-17-VQLA}} & \multicolumn{3}{c}{\textbf{Cholec80-VQA}} \\
    & & \textbf{Acc.} & \textbf{Rec.} & \textbf{m-F}$_{1}$ & \textbf{Acc.} & \textbf{w-F}$_{1}$ & \textbf{m-F}$_{1}$ & \textbf{Acc.} & \textbf{Rec.} & \textbf{m-F}$_{1}$ \\
    \hline
    \hline
    \rowcolor{lightgray!50}\multicolumn{11}{l}{\textit{Classification-based Models: Representative Research}} \\
    \hline
    {MFB} \cite{mfb} & LSTM & $52.4$ & $42.1$ & $36.2$ & $-$ & $-$ & $-$ & $84.1$ & $53.0$ & $45.9$ \\
    {MFH} \cite{mfh} & LSTM & $58.8$ & \colorbox{yellow!30}{$48.4$} & $42.2$ & $-$ & $-$ & $-$ & $87.5$ & $59.0$ & $55.7$ \\
    {VisualBert} \cite{visualbert} & BERT & $61.4$ & $42.8$ & $37.5$ & $39.0$ & $-$ & $31.7$ & $90.1$ & $62.9$ & $63.0$ \\
    {VisualBert ResMLP} \cite{sugvqa} & BERT & $61.9$ & $40.8$ & $35.8$ & $42.0$ & $-$ & \colorbox{green!15}{$33.2$} & $90.1$ & ${65.7}$ & $65.9$ \\
    {Surgical-VQLA} \cite{surgicalvqal} & BERT & \colorbox{green!15}{$66.6$} & $-$ & $36.1$ & \colorbox{green!15}{$45.8$} & $-$ & $24.5$ & $-$ & $-$ & $-$ \\ 
    {SurgicalGPT} \cite{suggpt} & GPT-2 & $66.1$ & $44.6$ & \colorbox{yellow!30}{$45.4$} & $-$ & $-$ & $-$ & $94.3$ & $73.4$ & $74.4$ \\
    {SSG-VQA-Net} \cite{ssgvqa} & Trans. & $-$ & $-$ & $-$ & $-$ & $-$ & $-$ & $90.6$ & $64.4$ & $63.7$ \\
    \hline\hline
    \rowcolor{lightgray!50}\multicolumn{11}{l}{\textit{Generation-based Models: Zero-Shot MLLMs}} \\
    \hline
    Med-Flamingo \cite{medflamingo} & Llama$_\text{7B}$ & $0.0$ & $0.0$ & $0.0$ & $0.0$ & $0.0$ & $0.0$ & $0.3$ & $0.4$ & $0.7$\\
    LLaVA-Med-v1.5 \cite{llavamed} & Mistral$_\text{7B}$ & $4.8$ & $14.4$ & $9.3$ &$2.8$ & $3.2$ & $7.0$ & $16.9$ & $14.4$ & $7.3$ \\
    HuatuoGPT-Vision \cite{huatuogptvision} & Qwen2$_\text{7B}$ & $7.3$ & $15.5$ & $11.0$ & $0.6$ & $1.2$ & $2.0$ & $30.8$ & $16.1$ & $10.5$ \\
    \hline\hline
    \rowcolor{lightgray!50}\multicolumn{11}{l}{\textit{Generation-based Models: LoRA-Tuned MLLMs}} \\
    \hline
    LLaVA-Med-v1.5 \cite{llavamed} & Mistral$_\text{7B}$ & $62.9$ & $37.0$ & $38.0$ & $40.7$ & $33.3$ & $32.5$ & $92.9$ & $71.3$ & $70.4$ \\
    HuatuoGPT-Vision \cite{huatuogptvision} & Qwen2$_\text{7B}$ & $61.5$ & $39.7$ & $32.6$ & $\underline{44.1}$ & \colorbox{green!15}{$\underline{38.8}$} & \colorbox{yellow!30}{$\bm{39.3}$} & $93.2$ & $71.8$ & $72.1$ \\
    BLIP-3 \cite{blip3} & Phi-3$_\text{3.8B}$ & $\underline{65.7}$ & $\underline{41.0}$ & $\underline{41.2}$ & $35.8$ & $32.9$ & $32.6$ & \colorbox{green!15}{$\underline{94.4}$} & \colorbox{green!15}{$\underline{75.7}$} & \colorbox{green!15}{$\underline{76.9}$}  \\
    \cdashline{1-11}[2pt/2pt]
    {\textsc{S$^2$Can} (Ours)} & Phi-3$_\text{3.8B}$ & \colorbox{yellow!30}{$\bm{69.6}$} & \colorbox{green!15}{$\bm{47.5}$} & \colorbox{green!15}{$\bm{42.7}$} & \colorbox{yellow!30}{$\bm{46.6}$} & \colorbox{yellow!30}{$\bm{42.8}$} & \colorbox{green!15}{$\underline{33.2}$} & \colorbox{yellow!30}{$\bm{94.7}$} & \colorbox{yellow!30}{$\bm{77.9}$} & \colorbox{yellow!30}{$\bm{77.9}$} \\
    \hline
    \end{tabular}}
    \caption{Experimental results on three surgical VQA datasets. The best results among MLLMs are highlighted in \textbf{boldface} and the second best are \underline{underlined}, while the global ones are marked with \colorbox{yellow!30}{scores} and \colorbox{green!15}{scores}. Results of classification-based models are cited from the literature. Otherwise, we implemented them based on their released repositories. Trans. denotes Transformer.}
    \label{table: all_results}
    \vspace{-10pt}
\end{table*}

%% file: section/experiment.tex
\begin{table}[h]
    \centering
    \resizebox{\linewidth}{!}{
    \begin{tabular}{l|cc|c|cc}
    \hline
    \multirow{2}{*}{\textbf{Stat.}}& \multicolumn{2}{c|}{\textbf{EndoVis-18-VQA}} & \textbf{EndoVis-17-VQLA} & \multicolumn{2}{c}{\textbf{Cholec80-VQA}} \\
    & \textbf{Train} & \textbf{Test} & \textbf{Test} & \textbf{Train} & \textbf{Test} \\\hline \hline
    \#Video & $11$ & $3$ & $5$ & $32$ & $8$\\
    \#Frame & $1,560$ & $447$ & $97$ & $17,043$ & $4,548$ \\
    \#QA & $9,014$ & $2,769$ & $472$ & $34,086$ & $9,096$ \\
    \#QA/F & $5.8$ & $6.2$ & $4.9$ & $2.0$ & $2.0$\\
    \#M/F & $4.0$ & $4.5$ & $4.2$ & $1.9$ & $1.9$ \\\cdashline{1-6}[1pt/1pt]
    \#A/Q & \multicolumn{2}{c|}{$3.6$} & $3.0$ & \multicolumn{2}{c}{$2.1$} \\
    \#Label & \multicolumn{2}{c|}{$18$} & $12$ & \multicolumn{2}{c}{$13$} \\
    Q Type & \multicolumn{2}{c|}{Action/Location} & Action/Location & \multicolumn{2}{c}{Binary/Count/Action} \\
    \hline
    \end{tabular}}
    \caption{The statistics of each dataset. \#M/F represents memory entries per frame and \#A/Q denotes the number of candidates per question.}
    \label{table: data_stat}
    \vspace{-10pt}
\end{table}

\begin{table*}[t]
    \centering
    \resizebox{\textwidth}{!}{
    \begin{tabular}{l|cc|ccc|ccc|ccc}
    \hline
    \multirow{2}{*}{\textbf{Model}} & \multicolumn{2}{c|}{\textbf{Memory Type}} & \multicolumn{3}{c|}{\textbf{EndoVis-18-VQA}} & \multicolumn{3}{c|}{\textbf{EndoVis-17-VQLA}} & \multicolumn{3}{c}{\textbf{Cholec80-VQA}} \\
    & $\mathcal{DM}$ (\ref{direct_memory}) & $\mathcal{IM}$ (\ref{indirect_memory}) & \textbf{Acc.} & \textbf{Rec.} & \textbf{m-F}$_{1}$ & \textbf{Acc.} & \textbf{Rec.} & \textbf{m-F}$_{1}$ & \textbf{Acc.} & \textbf{Rec.} & \textbf{m-F}$_{1}$ \\ 
    \hline\hline
    \textsc{S$^2$Can} & \cmark & \cmark & $69.6$ & $47.5$ & $42.7$ & $46.6$ & $40.1$ & $33.2$ & $94.7$ & $77.9$ & $77.9$ \\
    \textsc{S$^2$Can} \textit{w/o} $\mathcal{M}$ & \xmark & \xmark & $65.7$ & $41.0$ & $41.2$ & $35.8$ & $38.2$ & $32.6$ & $94.4$ & $75.7$ & $76.9$ \\
    \textsc{S$^2$Can} \textit{w/o} $\mathcal{DM}$ & \xmark & \cmark & $68.0$ & $41.9$ & $35.6$ & $46.0$ & $46.4$ & $30.2$ & $94.7$ & $77.8$ & $76.7$ \\
    \textsc{S$^2$Can} \textit{w/o} $\mathcal{IM}$ & \cmark & \xmark & $67.2$ & $41.3$ & $35.2$ & $44.9$ & $45.9$ & $36.3$ & $94.2$ & $73.1$ & $73.6$ \\
    \hline
    \end{tabular}}
    \caption{Overall ablation results on three datasets. \cmark denotes that the component is used, while \xmark denotes that it is removed.}
    \label{table: ablation_results}
    \vspace{-10pt}
\end{table*}

\section{Experiments}
\subsection{Datasets and Evaluation Metrics}
\textbf{Datasets.} To evaluate our models, we use three surgical VQA datasets, including EndoVis-18-VQA, EndoVis-17-VQLA, and Cholec80-VQA:
\begin{itemize}
    \item \textbf{EndoVis-18-VQA:} This dataset is derived from the MICCAI Endoscopic Vision Challenge 2018 dataset \cite{allan20202018roboticscenesegmentation}, consisting of 11,783 QA pairs from 14 videos. 
    \item \textbf{EndoVis-17-VQAL:} Derived from the MICCAI Endoscopic Vision Challenge 2017 dataset \cite{allan20192017roboticinstrumentsegmentation}, this dataset is annotated using the same process of the EndoVis-18-VQA with bounding box annotations. It contains 472 QA pairs from 5 videos and is only used for evaluating the generalizability of models. This dataset share the same labels of the EndoVis-18-VQA dataset.
    \item \textbf{Cholec80-VQA:} Annotated from the Cholec80 dataset \cite{cholec80}, this dataset includes 43,182 QA pairs from 40 videos.
\end{itemize}
Detailed statistics for these datasets are presented in Table \ref{table: data_stat}. Other details are provided in the Supplementary Material. \vspace{\submargin}

\noindent\textbf{Evaluation Metrics.} We use Accuracy (\textbf{Acc.}), Recall (\textbf{Rec.}), and macro-F$_1$ (\textbf{m-F}$_1$) as evaluation metrics. In addition, since the size of the EndoVis-17-VQAL dataset is relatively small, we use a weighted-F$_1$ (\textbf{w-F}$_1$) for better evaluation.

\begin{figure}
    \centering
    \includegraphics[width=1.0\linewidth]{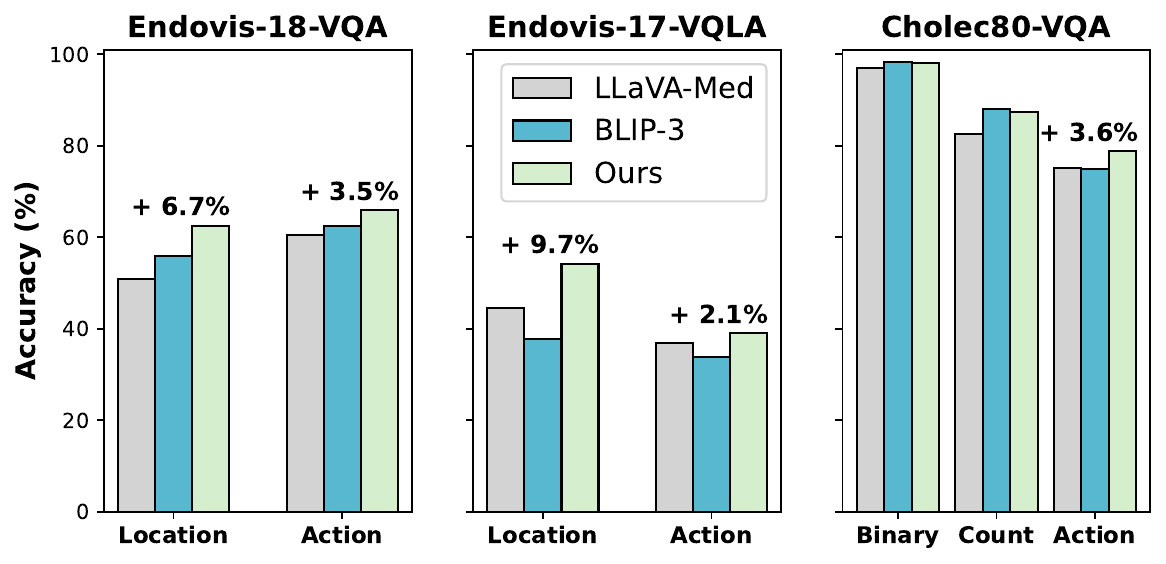}
    \caption{Performance of LLaVA-Med-v1.5, BLIP-3, and our \textsc{S$^2$Can} on different question types (e.g., Action or Location).}
    \label{fig:per_type}
\end{figure}

\subsection{Baselines}
We compare our \textsc{S$^2$Can} against SoTA baselines and MLLMs. The SoTA baselines include: \textbf{MFB} \cite{mfb} and \textbf{MFH} \cite{mfh}, which apply Factorized Bilinear and High-Order Pooling for cross-modal fusion, respectively; \textbf{VisualBert} \cite{visualbert}, a BERT-based model \cite{bert} pretrained with visually grounded objectives; \textbf{VisualBert ResMLP} \cite{sugvqa}, which incorporates a residual MLP module for surgical VQA; \textbf{Surgical-VQLA} \cite{surgicalvqal}, which answers visual questions and localize the corresponding objects; \textbf{SurgicalGPT} \cite{suggpt}, which leverages GPT-2’s \cite{gpt2} language priors for enhanced question understanding; and \textbf{SSG-VQA-Net} \cite{ssgvqa}, which incorporates objects and scene graph for improved VQA. We also compare three medical MLLM baselines: \textbf{Med-Flamingo} \cite{medflamingo}, based on Flamingo \cite{flamingo}; \textbf{LLaVA-Med-v1.5} \cite{llavamed}, which uses Llama-7B \cite{llama} or Mistral-7B \cite{mistral} for visually grounded medical text generation; and \textbf{HuatuoGPT-Vision} \cite{huatuogptvision}, built on Qwen2-7B \cite{qwen2}. Additionally, we evaluate a general-purpose MLLM, \textbf{BLIP-3} \cite{blip3}, a model with 4.59 billion parameters, based on Phi-3$_\text{3.8B}$ \cite{abdin2024phi3technicalreporthighly} and ViT-H/14-378 released by \cite{dfn_vit}, which also serves as the backbone for \textsc{S$^2$Can}. Zero-shot and LoRA settings are adopted to compare these MLLMs.

\subsection{Implementation Details}
The backbone of \textsc{S$^2$Can} is BLIP-3. In our approach, the visual encoder is kept frozen while the $\mathtt{Resampler}$ is fine-tuned. We utilize LoRA \cite{lora} to tune all linear layers of the language model, with $r=8$ and $\alpha=32$, and set the dropout rate to 0.1. The models are trained for 15 and 5 epochs on the EndoVis-18-VQA and Cholec80-VQA datasets, respectively, with a batch size of 32. The learning rate is set to $2e^{-5}$. The number of hints $K$ is set to 2 for all datasets, which is based on our preliminary experiments with $K\in \{2, 3\}$ and data statistics in Table \ref{table: data_stat} (\#A/Q). The number of indirect memory $M$ is set to 3 and 1 for these two datasets, respectively. Regarding the EndoVis-17-VQLA dataset, it is used to evaluate models trained on the EndoVis-18-VQA dataset. Note that during training, we randomly sample this number from $[1, M]$ to avoid over-fitting. The frequency $N$ of keeping a QA pair as indirect memory is set to $500$ for all datasets.

%% file: section/result.tex
\section{Results and Analysis}
\subsection{Quantitative Analysis}
\textbf{Overall Analysis.} Table \ref{table: all_results} presents the experimental results of representative SoTA methods, MLLM baselines, and our \textsc{S$^2$Can}. Our model achieves the highest accuracy, with significant improvements across the EndoVis-18-VQA (+3.9\%), EndoVis-17-VQLA (+2.5\%), and Cholec80-VQA (+0.3\%) datasets among MLLMs, demonstrating its robustness across different scenes. In the upper part of the table, we observe that small specialists (e.g., Surgical-VQLA and SurgicalGPT) leveraging pretrained vision or language models generally outperform those without pretrained models (e.g., MFB and MFH). When compared to LoRA-tuned MLLMs, these specialists remain competitive, particularly on the EndoVis-18-VQA and Cholec80-VQA datasets. Furthermore, when transferring knowledge from EndoVis-18-VQA to EndoVis-17-VQLA, \textsc{S$^2$Can} yields the best performance, further highlighting the robustness of our model.\vspace{\submargin}

\noindent\textbf{Analysis of Sub-Type Performance.} We analyze the performance of the models, LLaVA-Med-v1.5, BLIP-3 and \textsc{S$^2$Can}, across different question types. The results, shown in Table \ref{fig:per_type}, indicate that on the EndoVis-18-VQA dataset, our model consistently outperforms BLIP-3 across all question types, with a notable +6.7\% improvement on Location questions and +3.5\% on Action questions. Similarly, on the EndoVis-17-VQLA dataset, \textsc{S$^2$Can} surpasses LLaVA-Med-v1.5 with a 9.7\% improvement on Location questions and a 2.1\% improvement on Action questions. Additionally, \textsc{S$^2$Can} achieves a 3.8\% gain on Action questions while maintaining comparable performance on the other question types. We also observe that misclassified actions on the EndoVis-18-VQA dataset are often labeled as \textit{Idle}, likely due to the higher prevalence of \textit{Idle} instances compared to other actions. Moreover, the model occasionally confuses \textit{left-bottom} with \textit{left-top} and \textit{right-bottom} with \textit{right-top}, as these pairs share common prefixes (i.e., \textit{left-} and \textit{right-}).\vspace{\submargin}
\input{section/figures_tables/case_study}

\noindent\textbf{Analysis of Zero-shot Setting.} We analyze the performance of three medical MLLMs: Med-Flamingo, LLaVA-Med-v1.5, and HuatuoGPT-Vision. The experimental results are presented in the middle section of Table \ref{table: all_results}. These models were primarily trained using image-text pairs from medical domains. Among the three, HuatuoGPT-Vision delivers the best overall performance. LLaVA-Med-v1.5 shows results comparable to HuatuoGPT-Vision on the EndoVis-18-VQA and EndoVis-17-VQLA datasets but exhibits significantly lower performance on the Cholec80-VQA dataset. In contrast, Med-Flamingo struggles, being able to answer only a subset of binary questions, thus demonstrating the weakest performance. Overall, the performance of all three models is relatively low, reflecting their limited knowledge of surgery. This subpar performance can be attributed to a domain shift: mainstream MLLMs are predominantly trained on radiology datasets, and directly applying them to surgical tasks results in poor performance on surgical VQA samples. Our findings suggest that when applying MLLMs to the surgical domain, specific and rich in-domain resources are essential for achieving good performance. \vspace{\submargin}

\noindent\textbf{Analysis of LoRA Setting.}\label{lora} In addition to the medical MLLMs discussed above, we include a general-purpose MLLM, BLIP-3, for comparison. Given the low performance of Med-Flamingo, we exclude it for comparison. Our analysis now covers three MLLM baselines in total. According to the experimental results in the lower part of Table \ref{table: all_results}, BLIP-3 achieves the best overall performance among baselines, particularly on the EndoVis-18-VQA and Cholec80-VQA datasets. We attribute these improvements primarily to its more advanced visual encoder: while both LLaVA-Med-v1.5 and HuatuoGPT-Vision use CLIP-ViT-L/14-336 \cite{clip} as their visual encoder, BLIP-3 employs ViT-H/14-378. The improvement in language models may also contribute, despite BLIP-3's Phi-3 being nearly half the size of Mistral and Qwen. Based on these findings, we adopt BLIP-3 as the backbone of \textsc{S$^2$Can} for further comparisons. Our model significantly outperforms BLIP-3, delivering higher accuracy and better F$_1$ scores. This indicates that {augmenting contextual information in surgical scenes enhances the model's ability to understand and align visual features with the questions}, leading to improved performance.

\subsection{Qualitative Analysis}
\textbf{Case Study.} We conduct a case study by comparing the answers produced by \textsc{S$^2$Can} and other baselines. In addition to the three LoRA-tuned MLLMs mentioned above, we include two closed-source MLLMs, i.e., Claude-3.5 Sonnet \cite{claude} and GPT-4o \cite{gpt4o}, for comparison. For each dataset, we select two VQA samples and we list these cases in Table \ref{table: case_study}. Memory information generated by \textsc{S$^2$Can} is placed in the second column for a better understanding. As we can see from these samples, all open-sourced models produce fairly good predictions on simple questions, e.g., Case (e): \textit{Is irrigator used in calot triangle dissection?} {This is because these simple questions are detailed and usually contain rich visual objects and attributes}, e.g., image phase \textit{calot triangle dissection} and tool \textit{irrigator}. Thus, all the models can easily answer it. However, models' performance drops when a challenge question is asked, e.g., Case (f): \textit{What is the phase of the image?} Compared with binary questions with readily informed visual information, a model will need to access the overall condition of an image before answering the question, and {memory information could serve as a good source of context for the models.} Regarding two closed-source MLLMs, both models can recognize objects and attributes in surgical scenes, and provide relevant answers. After reading their predictions, we find Claude 3.5 provides more relevant outputs than GPT-4o. However, since we simply feed images and questions without other instruction, their performance could be further improved by refining prompts.\vspace{\submargin}

\begin{figure}[t]
    \centering
    \includegraphics[width=1.0\linewidth]{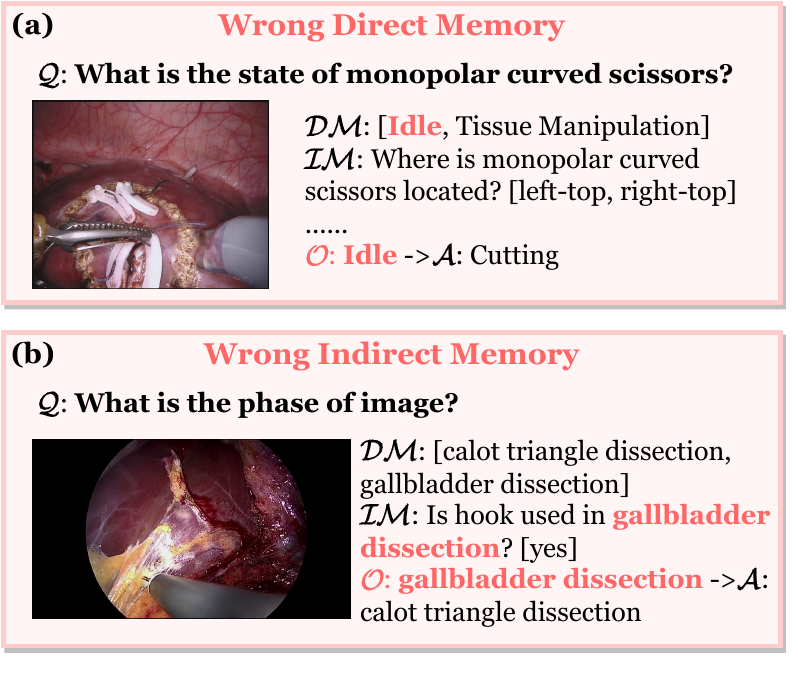}
    \caption{Two major causes of errors produced by \textsc{S$^2$Can}, i.e., wrong direct memory and wrong indirect memory. Case (a) and (b) are selected from the EndoVis-18-VQA and Cholec80-VQA datasets.}
    \label{fig:error_case}
\end{figure}

\noindent\textbf{Error Analysis.} To provide deeper insights, we present an error analysis in Figure \ref{fig:error_case}, identifying two major causes of errors produced by \textsc{S$^2$Can}: wrong indirect memory and wrong direct memory. In case (a), the model generates direct memory that includes two hints, i.e., \textit{Idle} and \textit{Tissue Manipulation}. However, the absence of the correct answer \textit{Cutting} leads the model to select the high-frequency but incorrect prediction \textit{Idle}. We believe that improving the design of hint structures, especially those with similar visual contexts, could help mitigate this issue. In case (b), where wrong indirect memory is the primary cause, the model directly copies an answer from memory. Given that indirect memory often consists of long text sequences, producing high-quality ones remains a challenging yet crucial task. One potential solution is increasing the number of training samples to enhance memory generation.
\begin{table}[t]
    \centering
    \resizebox{\linewidth}{!}{
    \begin{tabular}{c|cc|cc|cc}
        \hline
        \multirow{2}{*}{$\bm{M}$} & \multicolumn{2}{c|}{\footnotesize\textbf{EndoVis-18-VQA}} & \multicolumn{2}{c|}{\footnotesize\textbf{EndoVis-17-VQLA}} & \multicolumn{2}{c}{\footnotesize\textbf{Cholec80-VQA}} \\
        & \textbf{Acc.} & \textbf{m-F}$_1$ & \textbf{Acc.} & \textbf{m-F}$_1$ & \textbf{Acc.} & \textbf{m-F}$_1$ \\\hline\hline
        $0$ & $65.7$ & $41.2$ &$35.8$ & $32.6$  & $94.4$ & $76.9$ \\
        $1$ & $69.4$ & $40.6$ & $46.7$ & $29.0$ &$94.7$ & $77.9$ \\
        $2$ & $70.0$ & $42.1$ & $46.7$ & $32.3$ &$\color{gray!50} 94.7$ & $\color{gray!50} 77.9$  \\
        $3$ & $69.6$ & $42.7$ & $46.6$ & $33.2$ & $\color{gray!50} 94.7$ & $\color{gray!50} 77.9$ \\\hline
    \end{tabular}
    }
    \caption{Experimental results of \textsc{S$^2$Can} on three datasets with different numbers of entries in the indirect memory.}
    \label{tab:im_entry}
    \vspace{-10pt}
\end{table}

\subsection{Ablation Analysis}\label{sec: analysis}
\textbf{Analysis of Memory.} We investigate the effect of two types of memory (i.e., \textit{Indirect Memory} and \textit{Direct Memory}) and list the results in Table \ref{table: ablation_results}. The removal of memory leads to a significant drop in performance, highlighting the importance of incorporating contextual information. When comparing the variant \textit{w/o} $\mathcal{IM}$ to the variant \textit{w/o} $\mathcal{DM}$, we observe a larger performance decline in the absence of indirect memory. This suggests that indirect memory plays a more critical role in providing contextual information, which is reasonable given that $\mathcal{IM}$ contains more comprehensive information than $\mathcal{DM}$. Furthermore, since both types of memory share overlapping objectives, their combination further enhances the model's performance. \vspace{\submargin}

\noindent\textbf{Analysis of Entry Number in Indirect Memory.} We directly prompt \textsc{S$^2$Can} with $[1, M]$ entries of indirect memory for analysis, and the results are listed in Table \ref{tab:im_entry}. Note that the maximum number of entries for the EndoVis-18-VQA, EndoVis-17-VQLA, and Cholec80-VQA datasets are $3$, $3$, and $1$, respectively. We observe that when increasing the number of indirect memory \textsc{S$^2$Can}, the performance first increases slightly and then drops a bit. This indicates that a specific and targeted indirect memory is sufficient to augment the context for surgical VQA.

%% file: section/figures_tables/case_study.tex
\begin{table*}[t]
    \centering
    \resizebox{\linewidth}{!}{
    \begin{tabular}{c|l|l|l}
    \hline
    \textbf{Image} & \textbf{Question $\mathcal{Q}$ \& Answer $\mathcal{A}$ \& Memory $\mathcal{M}=\{\mathcal{DM}, \mathcal{IM}\}$} & \textbf{Model} & \textbf{Prediction} \\\hline\hline
    \rowcolor{lightgray!50}\multicolumn{4}{l}{\textit{EndoVis-18-VQA}} \\\hline
    \multirow{6}{*}{\begin{tabular}{@{}c@{}}
        \includegraphics[height=\dimexpr 7\ht\strutbox\relax, width=\dimexpr 11\ht\strutbox\relax]{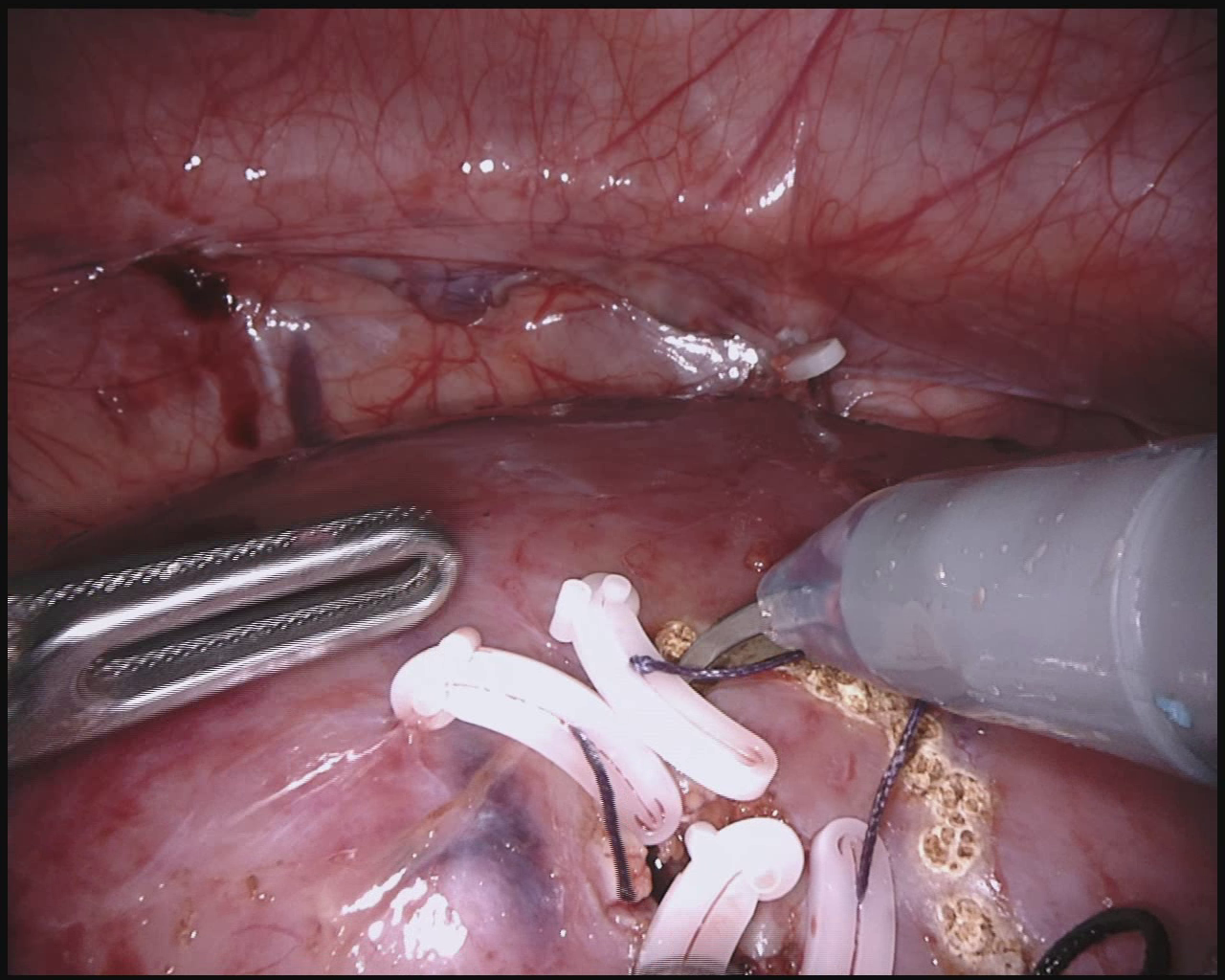} \\[-4px]
        (a) \\[-4px]
    \end{tabular}
    }
    &
    \multirow{6}{*}{
        \begin{tabular}{@{}p{0.5\textwidth}@{}}
        $\mathcal{Q}$: What is the state of prograsp forceps?\\
        $\mathcal{A}$: Tissue Manipulation \\
        $\mathcal{DM}$: [Idle, \hlomme{Tissue Manipulation}] \\
        $\mathcal{IM}$: \\
        Where is prograsp forceps located? [left-top, right-top] \\
        What is the state of bipolar forceps? [Idle, Tissue Manipulation]
        \end{tabular}
    } & Claude-3.5 Sonnet & \smark \dots manipulate tissue \dots \\
    & & GPT-4o & \smark \dots manipulate tissue \dots \\\cdashline{3-4}[1pt/1pt]
    & & LLaVA-Med-v1.5 &  \xmark Idle\\
    & & HuatuoGPT-Vision & \xmark Idle \\
    & & BLIP-3 & \xmark Idle \\\cdashline{3-4}[1pt/1pt]
    & & {\textsc{S$^2$Can} (Ours)} & \cmark {Tissue Manipulation} \\\hline
    \multirow{6}{*}{\begin{tabular}{@{}c@{}}
        \includegraphics[height=\dimexpr 7\ht\strutbox\relax, width=\dimexpr 11\ht\strutbox\relax]{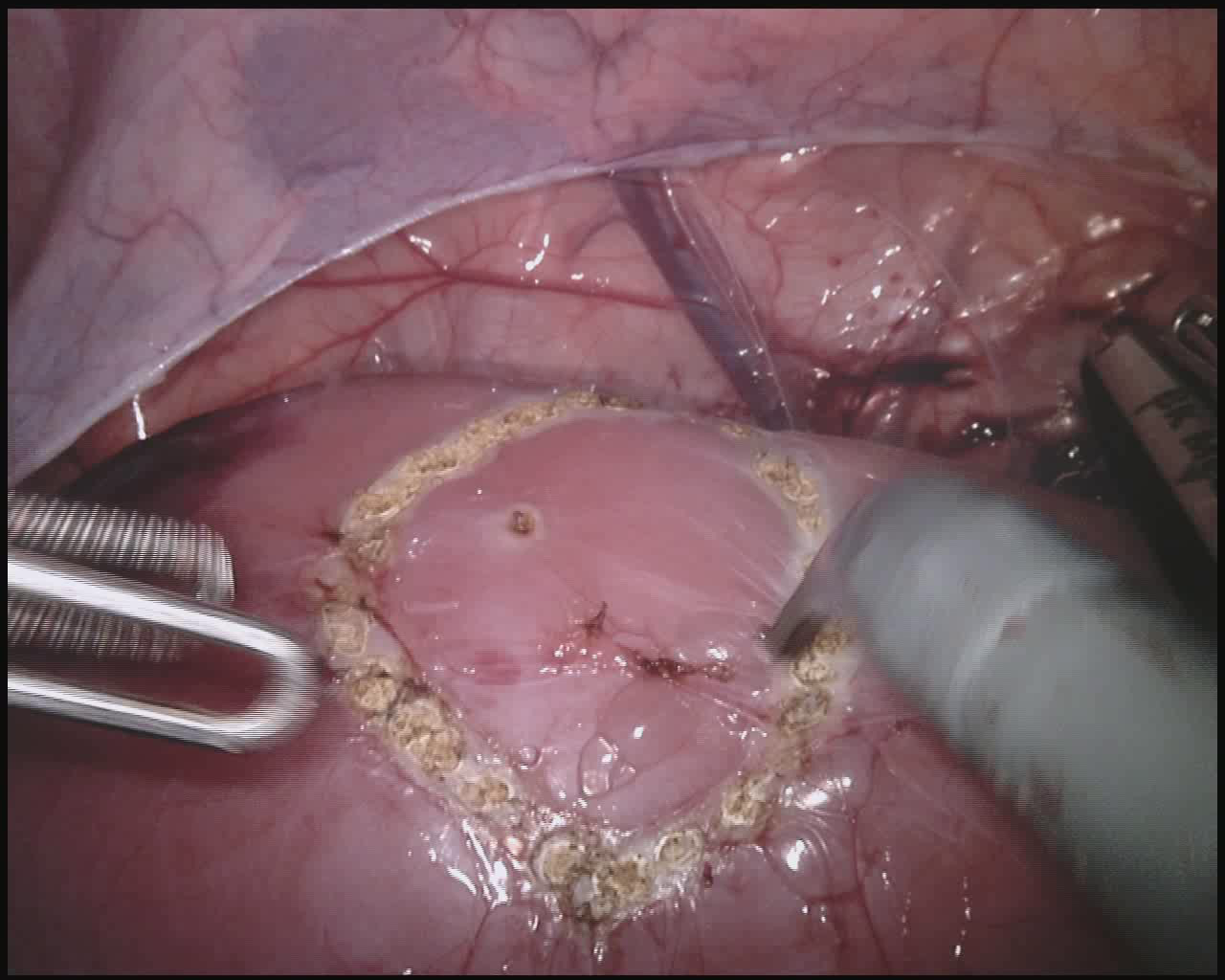}
        \\[-4px]
        (b) \\[-4px]
    \end{tabular}
    }
    & 
    \multirow{6}{*}{
        \begin{tabular}{@{}p{0.5\textwidth}@{}}
        $\mathcal{Q}$: Where is bipolar forceps located? $\mathcal{A}$: left-bottom \\
        $\mathcal{DM}$: [left-top, \hlomme{left-bottom}] \\
        $\mathcal{IM}$: \\
        What is the state of bipolar forceps? [Idle, Tissue Manipulation] \\
        Where is monopolar curved scissors located? [left-top, right-bottom]
        \end{tabular}
    } & Claude-3.5 Sonnet & \smark \dots left side \dots \\
    & & GPT-4o & \smark \dots left-hand side \dots \\\cdashline{3-4}[1pt/1pt]
    & & LLaVA-Med-v1.5 & \xmark left-top \\
    & & HuatuoGPT-Vision & \cmark {left-bottom} \\
    & & BLIP-3 & \xmark left-top \\ \cdashline{3-4}[1pt/1pt]
    & & {\textsc{S$^2$Can} (Ours)} &\cmark {left-bottom} \\\hline\hline
    \rowcolor{lightgray!50}\multicolumn{4}{l}{\textit{EndoVis-17-VQAL}} \\\hline
    \multirow{6}{*}{
    \begin{tabular}{@{}c@{}}
        \includegraphics[height=\dimexpr 7\ht\strutbox\relax, width=\dimexpr 11\ht\strutbox\relax]{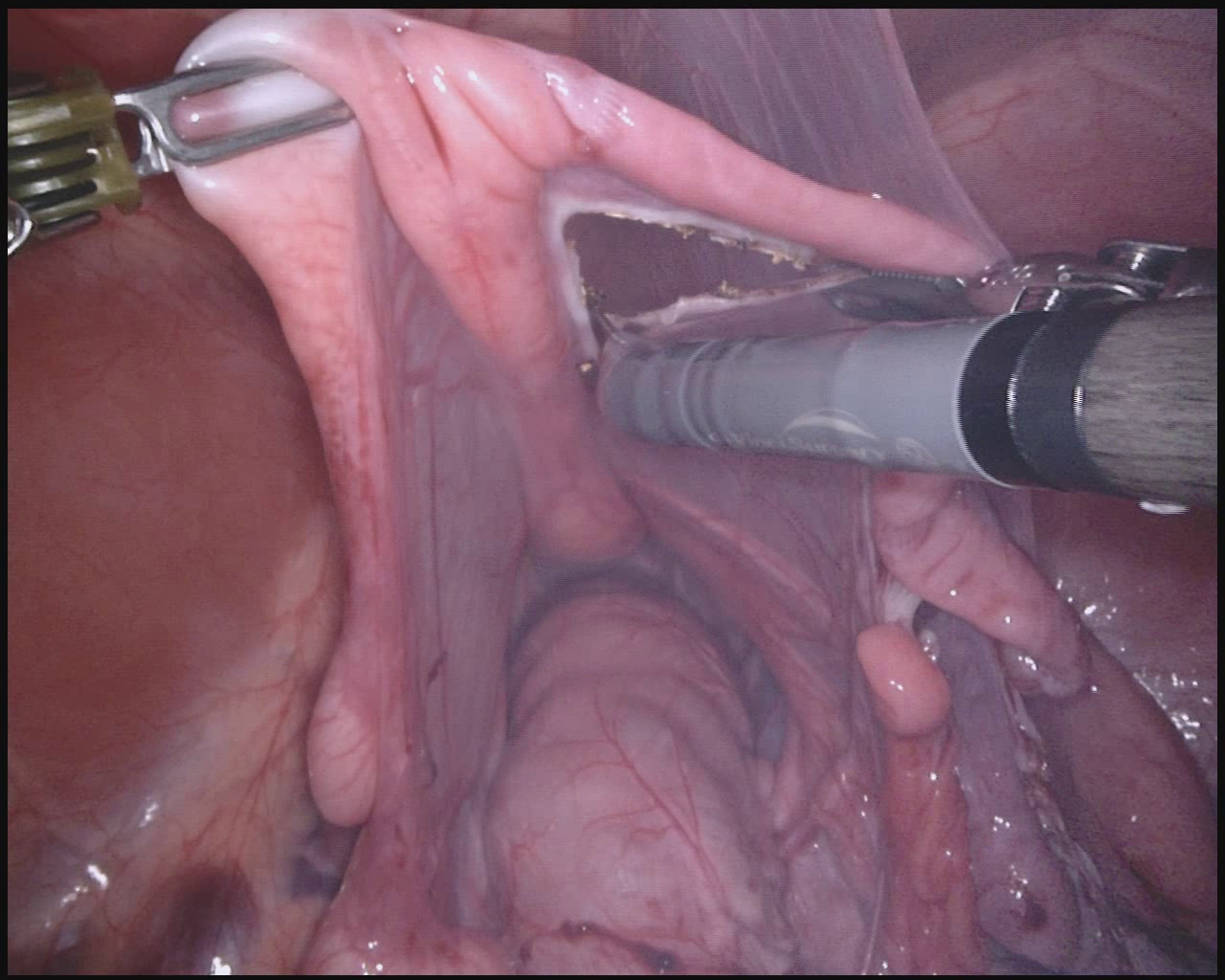}
        \\[-4px]
        (c) \\[-4px]
    \end{tabular}
    } & 
    \multirow{6}{*}{
        \begin{tabular}{@{}p{0.5\textwidth}@{}}
        $\mathcal{Q}$: Where is monopolar curved scissors located? \\
        $\mathcal{A}$: right-top $\mathcal{DM}$: [left-top, \hlomme{right-top}]  \\
        $\mathcal{IM}$: \\
        What is the state of monopolar curved scissors? [Idle, Tissue Manipulation]\\
        Where is bipolar forceps located? [left-top, left-bottom]
        \end{tabular}
    } & Claude-3.5 Sonnet & \xmark ... the center ... \\
    & & GPT-4o & \smark ... right side ... \\\cdashline{3-4}[1pt/1pt]
    & & LLaVA-Med-v1.5 & \cmark right-top \\
    & & HuatuoGPT-Vision & \cmark right-top \\
    & & BLIP-3 & \xmark right-top \\ \cdashline{3-4}[1pt/1pt]
    & & {\textsc{S$^2$Can} (Ours)} & \cmark right-top \\\hline
    \multirow{6}{*}{
        \begin{tabular}{@{}c@{}}
            \includegraphics[height=\dimexpr 7\ht\strutbox\relax, width=\dimexpr 11\ht\strutbox\relax]{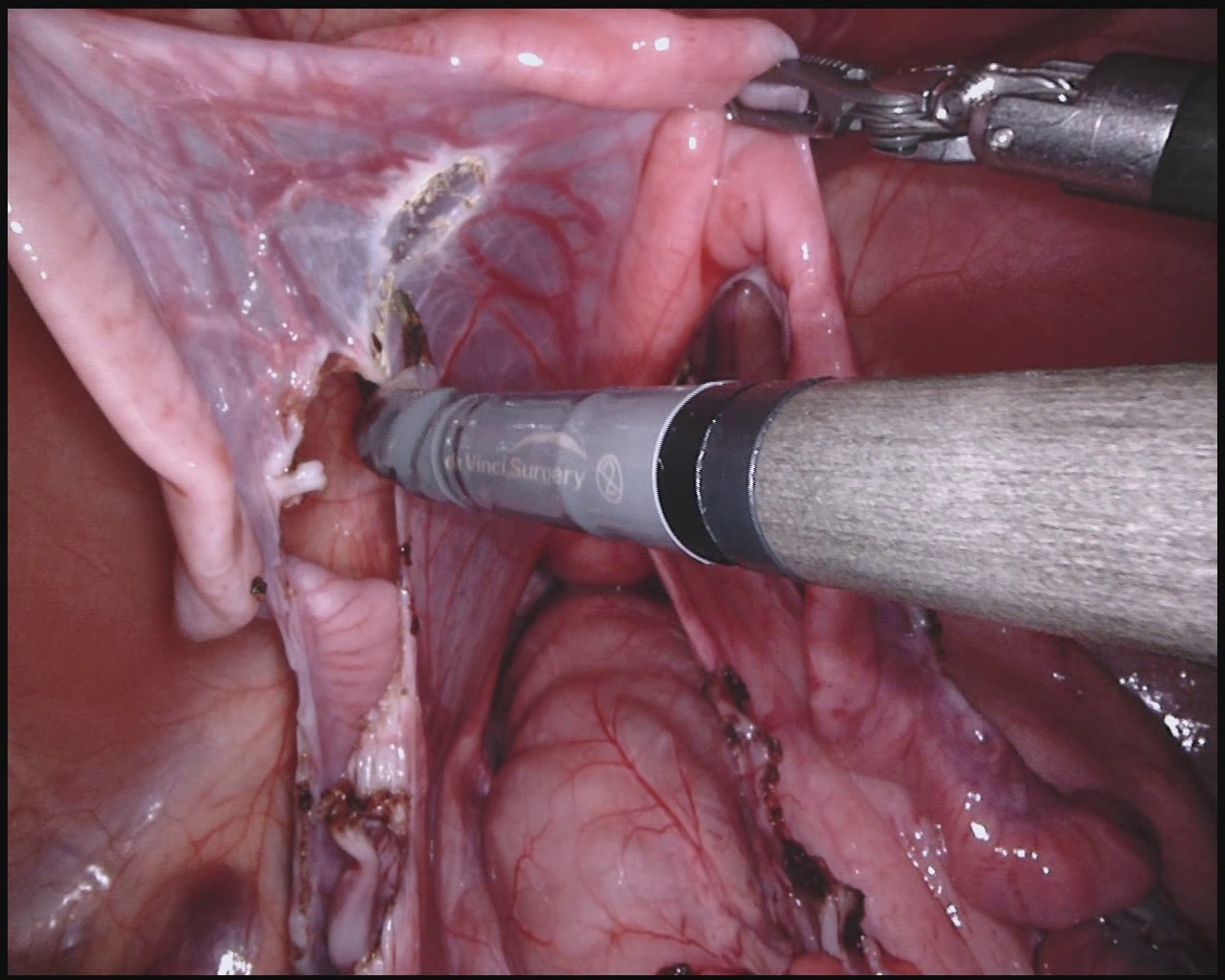}
            \\[-4px]
            (d) \\[-4px]
        \end{tabular}
        } & 
        \multirow{6}{*}{
            \begin{tabular}{@{}p{0.5\textwidth}@{}}
            $\mathcal{Q}$: What is the state of prograsp forceps? $\mathcal{A}$: Grasping \\
            $\mathcal{DM}$: [Idle, Tissue Manipulation] \\
            $\mathcal{IM}$: \\
            Where is prograsp forceps located? [left-top, right-top] \\
            What is the state of bipolar forceps? [Idle, Tissue Manipulation]\\
            \end{tabular}
        } & Claude-3.5 Sonnet & \cmark ... tissue manipulation ... \\
        & & GPT-4o & \smark ... manipulating tissue. \\ \cdashline{3-4}[1pt/1pt]
        & & LLaVA-Med-v1.5 & \xmark Tissue Manipulation \\
        & & HuatuoGPT-Vision & \xmark Tissue Manipulation \\
        & & BLIP-3 & \xmark Tissue Manipulation \\ \cdashline{3-4}[1pt/1pt]
        & & {\textsc{S$^2$Can} (Ours)} & \cmark Grasping \\\hline\hline
    \rowcolor{lightgray!50}\multicolumn{4}{l}{\textit{Cholec80-VQA}} \\\hline
    \multirow{6}{*}{
    \begin{tabular}{@{}c@{}}
        \includegraphics[height=\dimexpr 7\ht\strutbox\relax, width=\dimexpr 11\ht\strutbox\relax]{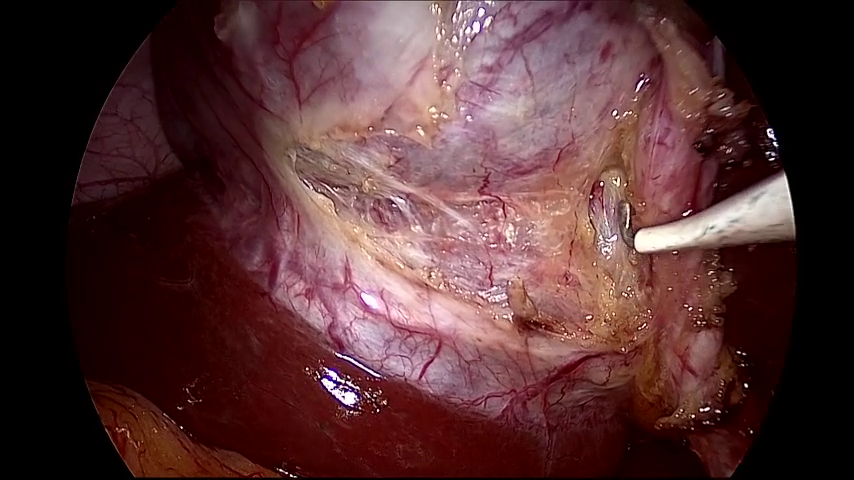}
        \\[-4px]
        (e) \\[-4px]
    \end{tabular}
    } & 
    \multirow{6}{*}{
        \begin{tabular}{@{}p{0.5\textwidth}@{}}
        $\mathcal{Q}$: What is the phase of image? \\
        $\mathcal{A}$: calot triangle dissection \\
        $\mathcal{DM}$: [\hlomme{calot triangle dissection}, gallbladder dissection] \\
        $\mathcal{IM}$: \\
        Is specimen bag used in \hlomme{calot triangle dissection}? [no] \\
        \end{tabular}
    } & Claude-3.5 Sonnet & \smark \dots dissection/surgical phase \\
    & & GPT-4o & \smark \dots dissection phase \dots \\\cdashline{3-4}[1pt/1pt]
    & & LLaVA-Med-v1.5 & \xmark gallbladder dissection \\
    & & HuatuoGPT-Vision & \xmark gallbladder dissection \\
    & & BLIP-3 & \cmark {calot triangle dissection} \\ \cdashline{3-4}[1pt/1pt]
    & & {\textsc{S$^2$Can} (Ours)} & \cmark {calot triangle dissection} \\\hline
    \multirow{6}{*}{
    \begin{tabular}{@{}c@{}}
        \includegraphics[height=\dimexpr 7\ht\strutbox\relax, width=\dimexpr 11\ht\strutbox\relax]{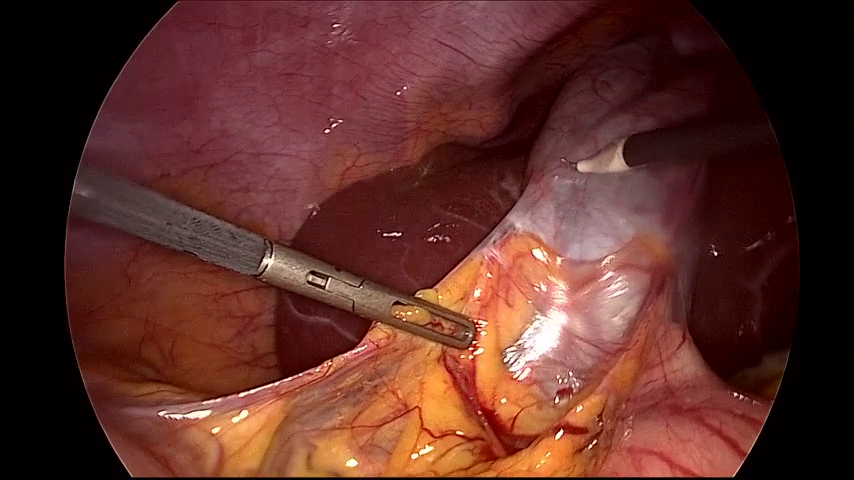}
        \\[-4px]
        (f) \\[-4px]
    \end{tabular}
    } & 
    \multirow{6}{*}{
        \begin{tabular}{@{}p{0.5\textwidth}@{}}
        $\mathcal{Q}$: Is irrigator used in calot triangle dissection? \\
        $\mathcal{A}$: no \\
        $\mathcal{DM}$: [NULL] \\
        $\mathcal{IM}$: \\
        Is {hook} used in calot triangle dissection? [{yes}] \\
        \end{tabular}
    } & Claude-3.5 Sonnet & \cmark no \dots \\
    & & GPT-4o & \xmark yes \dots \\\cdashline{3-4}[1pt/1pt]\cdashline{3-4}[1pt/1pt]
    & & LLaVA-Med-v1.5 & \cmark no \\
    & & HuatuoGPT-Vision & \cmark no \\
    & & BLIP-3 & \cmark no \\ \cdashline{3-4}[1pt/1pt]
    & & {\textsc{S$^2$Can} (Ours)} & \cmark no \\\hline
    \end{tabular}}
    \caption{Case study on three datasets comparing \textsc{S$^2$Can} with two closed-source MLLMs (Claude-3.5 and GPT-4o) and three open-source MLLMs. Only \textsc{S$^2$Can} uses memory as input, formatted as "Question [Hints]". A \cmark indicates a correct answer, a \xmark denotes an incorrect answer, and a \smark represents a partially correct prediction. Highlighted \hlomme{spans} indicate content closely related to the final predictions. Full outputs of Claude-3.5 Sonnet and GPT-4o are provided in Supplementary Material.}\label{table: case_study}
    \vspace{-10pt}
\end{table*}

%% file: section/conclusion.tex
\section{Conclusion}
In this paper, we tackled the task of Surgical VQA by introducing \textsc{S$^2$Can}, a simple yet effective memory-augmented MLLM. Our model operates autonomously, without relying on external resources, generating both direct and indirect memory through self-contained inquiry, based solely on the provided images and user questions. By reasoning over these memories, \textsc{S$^2$Can} significantly improves the comprehension of surgical scenes and questions. Experimental results with comprehensive analyses of three Surgical VQA datasets validate the effectiveness of our approach, demonstrating its potential to advance the field.

%% file: section/X_suppl.tex
\clearpage
\appendix
\maketitlesupplementary

\section{Prompt Templates of \textsc{S$^2$Can}}\label{appendix: prompt_template}
The prompt templates of \textsc{S$^2$Can} are listed in Table \ref{tab:prompt_template}, including VQA, Direct Memory Creation, Indirect Memory Creation, and Memory-Augmented VQA. Specifically, the prompts for memory creation are as follows:
\begin{itemize}
    \item Direct Memory $\mathcal{P}_{\mathcal{DM}}$: \texttt{Generate some hints for the answer.}
    \item Indirect Memory $\mathcal{P}_{\mathcal{IM}}$: \texttt{Generate a list of relevant questions with hints based on the provided surgical image.}
\end{itemize}

\section{Additional Implementation Details}\label{appendix: impl_details}
The model card of BLIP-3 is \href{https://huggingface.co/Salesforce/xgen-mm-phi3-mini-instruct-r-v1}{Salesforce/xgen-mm-phi3-mini-instruct-r-v1}. We access Claude-3.5 Sonnet and GPT-4o through the API of \href{https://poe.com/}{Poe}. The decoding length of indirect memory is set to 160 with a beam size of 3, while the length of direct memory is set to 12, 12, and 16 for three datasets using greedy decoding for the EndoVis-18-VQA, EndoVis-17-VQLA, and Cholec80-VQA datasets, respectively.

\section{Additional Data Details}\label{appendix: data_desc}
\subsection{Memory Annotation Details}
For the Cholec80-VQA dataset, we exclude binary questions during hint generation (\S\ref{direct_memory}) because they only correspond to yes/no answers. We also exclude questions that only have one answer for memory-augmented VQA.

\subsection{Questions and Answers of Datasets}
Questions of each dataset are listed in Table \ref{tab:dataset_qa_list}. In addition, their class labels are:
\begin{itemize}
    \item EndoVis-18-VQA and EndoVis-17-VQLA: {kidney}, {Idle}, {Grasping}, {Retraction}, {Tissue\_Manipulation}, {Tool\_Manipulation}, {Cutting}, {Cauterization}, {Suction}, {Looping}, {Suturing}, {Clipping}, {Staple}, {Ultrasound\_Sensing}, {left-top}, {right-top}, {left-bottom}, and {right-bottom}.
    \item Cholec80-VQA: {no}, {yes}, {0}, {1}, {2}, {3}, {calot triangle dissection}, {gallbladder dissection}, {clipping cutting}, {gallbladder retraction}, {cleaning coagulation}, {gallbladder packaging}, and {preparation}.
\end{itemize}

\begin{table*}[h]
\centering
\begin{tabular}{p{17cm}}
\hline
\rowcolor{lightgray!50}{\textit{EndoVis-18-VQA}} \\\hline
\textbf{Case (a):} What is the state of prograsp forceps? \\\cdashline{1-1}[1pt/1pt]
\textbf{Claude-3.5 Sonnet:} In this laparoscopic surgical image, the ProGrasp forceps appear to be in good working condition and are directly being used during the procedure. They are visible in the surgical field with the characteristic white grasping tips and being utilized to \smark \textcolor{red!50}{manipulate tissue}. \\\cdashline{1-1}[1pt/1pt]
\textbf{GPT-4o:} The progress forceps in the image are in an open state. The jaws of the instrument are not closed, indicating that they are ready to grasp or \smark \textcolor{red!50}{manipulate tissue}, but are currently not applying pressure or holding anything. \\
\hline
\textbf{Case (b):} Where is bipolar forceps located?  \\\cdashline{1-1}[1pt/1pt]
\textbf{Claude-3.5 Sonnet:} In this surgical image, the bipolar forceps can be seen on the \smark \textcolor{red!50}{left side} of the image, appearing as metallic/silver-colored surgical instruments with a distinctive ridged pattern. \\\cdashline{1-1}[1pt/1pt]
\textbf{GPT-4o:} The bipolar forceps are located on the \smark \textcolor{red!50}{left-hand side} of the image. They are the metallic instrument with two prongs visibly entering the surgical area. Bipolar forceps are commonly used in surgical procedures for grasping and coagulating tissue, often in minimally invasive or laparoscopic surgeries. \\
\hline\hline
\rowcolor{lightgray!50}{\textit{EndoVis-17-VQLA}} \\\hline
\textbf{Case (c):} Where is monopolar curved scissors located? \\\cdashline{1-1}[1pt/1pt]
\textbf{Claude-3.5 Sonnet:} This appears to be an endoscopic view during surgery where the monopolar curved scissors are located in the surgical field, being used to dissect or cut tissue. The image shows the curved scissors in what appears to be a laparoscopic view with pink/reddish tissue surrounding it. The metallic instrument is visible in the center of the image. \\\cdashline{1-1}[1pt/1pt]
\textbf{GPT-4o:} In the image, the monopolar curved scissors are located on the \smark \textcolor{red!50}{right side}, extending into the surgical field. The scissors are being used to cut tissue, and their curved blades can be seen making contact with a structure, likely in the abdominal cavity during a laparoscopic procedure. The instrument is connected to a robotic or laparoscopic system, as indicated by the handle and shaft visible in the image. \\\hline
\textbf{Case (d):} What is the state of prograsp forceps? \\\cdashline{1-1}[1pt/1pt]
\textbf{Claude-3.5 Sonnet:} The image shows a da Vinci surgical robot instrument, specifically the ProGrasp forceps, in use during a surgical procedure. The forceps appear to be functional and actively engaged in \smark \textcolor{red!50}{tissue manipulation}, as evidenced by their positioning within the surgical field. They are marked with the "Vinci Surgery" label visible on their shaft. \\\cdashline{1-1}[1pt/1pt]
\textbf{GPT-4o:} The prograsp forceps in the image are in a clamped or closed state, actively holding or \smark \textcolor{red!50}{manipulating tissue}. The forceps, likely part of a robotic surgical system (such as the da Vinci system), appear to be gripping a portion of the tissue, which is being dissected or manipulated during minimally invasive surgery. The tissue structure seems to be under tension, suggesting that the forceps are being used to maintain traction or provide exposure to a particular area for the surgical procedure. \\\hline\hline
\rowcolor{lightgray!50}{\textit{Cholec80-VQA}} \\\hline
\textbf{Case (e):} What is the phase of image? \\\cdashline{1-1}[1pt/1pt]
\textbf{Claude-3.5 Sonnet:} This appears to be a laparoscopic surgical image taken through an endoscope. Based the visible tissue characteristics, blood vessels, and the surgical instrument in view, this is likely during the \smark \textcolor{red!50}{dissection/surgical phase} of a laparoscopic procedure. The image shows vascularized tissue with a surgical instrument (likely a dissector or grasper) visible on the right side of the frame. \\\cdashline{1-1}[1pt/1pt]
\textbf{GPT-4o:} Given the appearance of the anatomy and the nature of the procedure, this may be part of the \smark \textcolor{red!50}{dissection phase} during a laparoscopic cholecystectomy (gallbladder removal surgery), where surgeons carefully separate tissues to isolate the cystic duct and artery. \\
\hline
\textbf{Case (f):} Is irrigator used in calot triangle dissection? \\\cdashline{1-1}[1pt/1pt]
\textbf{Claude-3.5 Sonnet:} \smark \textcolor{red!50}{No}, irrigation is typically not used during Calot's triangle dissection in laparoscopic cholecystectomy. \\\cdashline{1-1}[1pt/1pt]
\textbf{GPT-4o:} \smark \textcolor{red!50}{Yes}, an irrigator can be used during the dissection of the Calot's triangle (as in laparoscopic cholecystectomy), particularly for cleaning blood, bile, or other fluids to improve visibility of the dissection area. \\
\hline
\end{tabular}
\caption{Full responses of Claude-3.5 Sonnet and GPT-4o for the cases in Table \ref{table: case_study}. \smark \textcolor{red!50}{spans} are the partially correct predictions.}
\label{tab:dataset_qa_list}
\end{table*}

\begin{table*}[h!]
\centering
\begin{tabular}{c|p{12cm}}
\hline
\textbf{Purpose} &  \multicolumn{1}{c}{\textbf{Prompt Template}} \\ \hline
    \multirow{5}{*}{Vanilla VQA} & \texttt{<|user|>} \\
    & \texttt{<image>} \\
    & \texttt{\{Question\}<|end|>} \\
    & \texttt{<|assistant|>} \\
    & \texttt{\{Answer\}<|end|>} \\\hline
    \multirow{5}{*}{\makecell{Direct Memory\\Creation (\S\ref{direct_memory})}} & \texttt{<|user|>} \\
    & \texttt{<image>} \\
    & \texttt{\{Question\} Generate some hints for the answer.<|end|>} \\
    & \texttt{<|assistant|>} \\
    & \texttt{\{Hints\}<|end|>} \\\hline
    \multirow{7}{*}{\makecell{Indirect Memory\\Creation (\S\ref{indirect_memory})}} & \texttt{<|user|>} \\
    & \texttt{<image>} \\
    & \texttt{Generate a list of relevant questions with hints based on the provided surgical image.<|end|>} \\
    & \texttt{<|assistant|>} \\
    & \texttt{\{Question 1\} \{Hints\}} \\
    & \texttt{\{Question 2\} \{Hints\} \dots<|end|>} \\\hline
    \multirow{9}{*}{\makecell{Memory-Augmented\\VQA (\S\ref{mvqa})}} & \texttt{<|user|>} \\
    & \texttt{<image>} \\
    & \texttt{Memory:} \\
    & \texttt{\{Question 1\} \{Hints\}} \\
    & \texttt{\{Question 2\} \{Hints\} \dots} \\
    & \texttt{Question:} \\
    & \texttt{\{Question\} \{Hints\}<|end|>} \\
    & \texttt{<|assistant|>} \\
    & \texttt{\{Answer\}<|end|>} \\\hline
\end{tabular}
\caption{Prompt templates used for VQA, Direct Memory Creation, Indirect Memory Creation, and Memory-Augmented VQA.}
\label{tab:prompt_template}
\end{table*}

%% file: main.bbl
\begin{thebibliography}{58}
\providecommand{\natexlab}[1]{#1}
\providecommand{\url}[1]{\texttt{#1}}
\expandafter\ifx\csname urlstyle\endcsname\relax
  \providecommand{\doi}[1]{doi: #1}\else
  \providecommand{\doi}{doi: \begingroup \urlstyle{rm}\Url}\fi

\bibitem[Abdin et~al.(2024)Abdin, Jacobs, Awan, Aneja, Awadallah, Awadalla, Bach, Bahree, Bakhtiari, Bao, Behl, Benhaim, Bilenko, Bjorck, Bubeck, Cai, Cai, Mendes, Chen, Chaudhary, Chen, Chen, Chen, Chen, Chopra, Dai, Giorno, de~Rosa, Dixon, Eldan, Fragoso, Iter, Gao, Gao, Gao, Garg, Goswami, Gunasekar, Haider, Hao, Hewett, Huynh, Javaheripi, Jin, Kauffmann, Karampatziakis, Kim, Khademi, Kurilenko, Lee, Lee, Li, Li, Liang, Liden, Liu, Liu, Liu, Lin, Lin, Luo, Madan, Mazzola, Mitra, Modi, Nguyen, Norick, Patra, Perez-Becker, Portet, Pryzant, Qin, Radmilac, Rosset, Roy, Ruwase, Saarikivi, Saied, Salim, Santacroce, Shah, Shang, Sharma, Shukla, Song, Tanaka, Tupini, Wang, Wang, Wang, Wang, Ward, Wang, Witte, Wu, Wyatt, Xiao, Xu, Xu, Xu, Yadav, Yang, Yang, Yang, Yang, Yu, Yuan, Zhang, Zhang, Zhang, Zhang, Zhang, Zhang, Zhang, and Zhou]{abdin2024phi3technicalreporthighly}
Marah Abdin, Sam~Ade Jacobs, Ammar~Ahmad Awan, Jyoti Aneja, Ahmed Awadallah, Hany Awadalla, Nguyen Bach, Amit Bahree, Arash Bakhtiari, Jianmin Bao, Harkirat Behl, Alon Benhaim, Misha Bilenko, Johan Bjorck, Sébastien Bubeck, Qin Cai, Martin Cai, Caio César~Teodoro Mendes, Weizhu Chen, Vishrav Chaudhary, Dong Chen, Dongdong Chen, Yen-Chun Chen, Yi-Ling Chen, Parul Chopra, Xiyang Dai, Allie~Del Giorno, Gustavo de Rosa, Matthew Dixon, Ronen Eldan, Victor Fragoso, Dan Iter, Mei Gao, Min Gao, Jianfeng Gao, Amit Garg, Abhishek Goswami, Suriya Gunasekar, Emman Haider, Junheng Hao, Russell~J. Hewett, Jamie Huynh, Mojan Javaheripi, Xin Jin, Piero Kauffmann, Nikos Karampatziakis, Dongwoo Kim, Mahoud Khademi, Lev Kurilenko, James~R. Lee, Yin~Tat Lee, Yuanzhi Li, Yunsheng Li, Chen Liang, Lars Liden, Ce Liu, Mengchen Liu, Weishung Liu, Eric Lin, Zeqi Lin, Chong Luo, Piyush Madan, Matt Mazzola, Arindam Mitra, Hardik Modi, Anh Nguyen, Brandon Norick, Barun Patra, Daniel Perez-Becker, Thomas Portet, Reid Pryzant, Heyang
  Qin, Marko Radmilac, Corby Rosset, Sambudha Roy, Olatunji Ruwase, Olli Saarikivi, Amin Saied, Adil Salim, Michael Santacroce, Shital Shah, Ning Shang, Hiteshi Sharma, Swadheen Shukla, Xia Song, Masahiro Tanaka, Andrea Tupini, Xin Wang, Lijuan Wang, Chunyu Wang, Yu Wang, Rachel Ward, Guanhua Wang, Philipp Witte, Haiping Wu, Michael Wyatt, Bin Xiao, Can Xu, Jiahang Xu, Weijian Xu, Sonali Yadav, Fan Yang, Jianwei Yang, Ziyi Yang, Yifan Yang, Donghan Yu, Lu Yuan, Chengruidong Zhang, Cyril Zhang, Jianwen Zhang, Li~Lyna Zhang, Yi Zhang, Yue Zhang, Yunan Zhang, and Xiren Zhou.
\newblock Phi-3 technical report: A highly capable language model locally on your phone, 2024.

\bibitem[Alayrac et~al.(2022)Alayrac, Donahue, Luc, Miech, Barr, Hasson, Lenc, Mensch, Millican, Reynolds, Ring, Rutherford, Cabi, Han, Gong, Samangooei, Monteiro, Menick, Borgeaud, Brock, Nematzadeh, Sharifzadeh, Bi\'{n}kowski, Barreira, Vinyals, Zisserman, and Simonyan]{flamingo}
Jean-Baptiste Alayrac, Jeff Donahue, Pauline Luc, Antoine Miech, Iain Barr, Yana Hasson, Karel Lenc, Arthur Mensch, Katherine Millican, Malcolm Reynolds, Roman Ring, Eliza Rutherford, Serkan Cabi, Tengda Han, Zhitao Gong, Sina Samangooei, Marianne Monteiro, Jacob~L Menick, Sebastian Borgeaud, Andy Brock, Aida Nematzadeh, Sahand Sharifzadeh, Miko\l~aj Bi\'{n}kowski, Ricardo Barreira, Oriol Vinyals, Andrew Zisserman, and Kar\'{e}n Simonyan.
\newblock Flamingo: a visual language model for few-shot learning.
\newblock In \emph{Advances in Neural Information Processing Systems}, pages 23716--23736. Curran Associates, Inc., 2022.

\bibitem[Alberti et~al.(2019)Alberti, Ling, Collins, and Reitter]{vqa_object2}
Chris Alberti, Jeffrey Ling, Michael Collins, and David Reitter.
\newblock Fusion of detected objects in text for visual question answering.
\newblock In \emph{Proceedings of the 2019 Conference on Empirical Methods in Natural Language Processing and the 9th International Joint Conference on Natural Language Processing (EMNLP-IJCNLP)}, pages 2131--2140, Hong Kong, China, 2019. Association for Computational Linguistics.

\bibitem[Allan et~al.(2019)Allan, Shvets, Kurmann, Zhang, Duggal, Su, Rieke, Laina, Kalavakonda, Bodenstedt, Herrera, Li, Iglovikov, Luo, Yang, Stoyanov, Maier-Hein, Speidel, and Azizian]{allan20192017roboticinstrumentsegmentation}
Max Allan, Alex Shvets, Thomas Kurmann, Zichen Zhang, Rahul Duggal, Yun-Hsuan Su, Nicola Rieke, Iro Laina, Niveditha Kalavakonda, Sebastian Bodenstedt, Luis Herrera, Wenqi Li, Vladimir Iglovikov, Huoling Luo, Jian Yang, Danail Stoyanov, Lena Maier-Hein, Stefanie Speidel, and Mahdi Azizian.
\newblock 2017 robotic instrument segmentation challenge, 2019.

\bibitem[Allan et~al.(2020)Allan, Kondo, Bodenstedt, Leger, Kadkhodamohammadi, Luengo, Fuentes, Flouty, Mohammed, Pedersen, Kori, Alex, Krishnamurthi, Rauber, Mendel, Palm, Bano, Saibro, Shih, Chiang, Zhuang, Yang, Iglovikov, Dobrenkii, Reddiboina, Reddy, Liu, Gao, Unberath, Kim, Kim, Kim, Kim, Lee, Ullah, Luna, Park, Azizian, Stoyanov, Maier-Hein, and Speidel]{allan20202018roboticscenesegmentation}
Max Allan, Satoshi Kondo, Sebastian Bodenstedt, Stefan Leger, Rahim Kadkhodamohammadi, Imanol Luengo, Felix Fuentes, Evangello Flouty, Ahmed Mohammed, Marius Pedersen, Avinash Kori, Varghese Alex, Ganapathy Krishnamurthi, David Rauber, Robert Mendel, Christoph Palm, Sophia Bano, Guinther Saibro, Chi-Sheng Shih, Hsun-An Chiang, Juntang Zhuang, Junlin Yang, Vladimir Iglovikov, Anton Dobrenkii, Madhu Reddiboina, Anubhav Reddy, Xingtong Liu, Cong Gao, Mathias Unberath, Myeonghyeon Kim, Chanho Kim, Chaewon Kim, Hyejin Kim, Gyeongmin Lee, Ihsan Ullah, Miguel Luna, Sang~Hyun Park, Mahdi Azizian, Danail Stoyanov, Lena Maier-Hein, and Stefanie Speidel.
\newblock 2018 robotic scene segmentation challenge, 2020.

\bibitem[Anthropic(2024)]{claude}
Anthropic.
\newblock The claude 3 model family: Opus, sonnet, haiku.
\newblock 2024.

\bibitem[Antol et~al.(2015)Antol, Agrawal, Lu, Mitchell, Batra, Zitnick, and Parikh]{vqa_iccv}
Stanislaw Antol, Aishwarya Agrawal, Jiasen Lu, Margaret Mitchell, Dhruv Batra, C.~Lawrence Zitnick, and Devi Parikh.
\newblock Vqa: Visual question answering.
\newblock In \emph{Proceedings of the IEEE International Conference on Computer Vision (ICCV)}, 2015.

\bibitem[Bae et~al.(2024)Bae, Kyung, Ryu, Cho, Lee, Kweon, Oh, Ji, Chang, Kim, et~al.]{ehrxqa}
Seongsu Bae, Daeun Kyung, Jaehee Ryu, Eunbyeol Cho, Gyubok Lee, Sunjun Kweon, Jungwoo Oh, Lei Ji, Eric Chang, Tackeun Kim, et~al.
\newblock Ehrxqa: A multi-modal question answering dataset for electronic health records with chest x-ray images.
\newblock \emph{Advances in Neural Information Processing Systems}, 36, 2024.

\bibitem[Bai et~al.(2023{\natexlab{a}})Bai, Bai, Yang, Wang, Tan, Wang, Lin, Zhou, and Zhou]{qwenvl}
Jinze Bai, Shuai Bai, Shusheng Yang, Shijie Wang, Sinan Tan, Peng Wang, Junyang Lin, Chang Zhou, and Jingren Zhou.
\newblock Qwen-vl: A versatile vision-language model for understanding, localization, text reading, and beyond, 2023{\natexlab{a}}.

\bibitem[Bai et~al.(2023{\natexlab{b}})Bai, Islam, Seenivasan, and Ren]{surgicalvqal}
Long Bai, Mobarakol Islam, Lalithkumar Seenivasan, and Hongliang Ren.
\newblock Surgical-vqla:transformer with gated vision-language embedding for visual question localized-answering in robotic surgery.
\newblock In \emph{2023 IEEE International Conference on Robotics and Automation (ICRA)}, pages 6859--6865, 2023{\natexlab{b}}.

\bibitem[Bai et~al.(2024)Bai, Wang, Islam, Seenivasan, Wang, and Ren]{surgicalvqal++}
Long Bai, Guankun Wang, Mobarakol Islam, Lalithkumar Seenivasan, An Wang, and Hongliang Ren.
\newblock Surgical-vqla++: Adversarial contrastive learning for calibrated robust visual question-localized answering in robotic surgery, 2024.

\bibitem[{Ben Abacha} et~al.(2019){Ben Abacha}, Hasan, Datla, Liu, Demner-Fushman, and M\"uller]{ImageCLEFVQA-Med2019}
Asma {Ben Abacha}, Sadid~A. Hasan, Vivek~V. Datla, Joey Liu, Dina Demner-Fushman, and Henning M\"uller.
\newblock Vqa-med: Overview of the medical visual question answering task at imageclef 2019.
\newblock In \emph{Working Notes of {CLEF} 2019}, Lugano, Switzerland, 2019. CEUR-WS.org.

\bibitem[Ben-younes et~al.(2017)Ben-younes, Cadene, Cord, and Thome]{mutan}
Hedi Ben-younes, Remi Cadene, Matthieu Cord, and Nicolas Thome.
\newblock Mutan: Multimodal tucker fusion for visual question answering.
\newblock In \emph{2017 IEEE International Conference on Computer Vision (ICCV)}, pages 2631--2639, 2017.

\bibitem[Ben-younes et~al.(2019)Ben-younes, Cadene, Thome, and Cord]{block}
Hedi Ben-younes, Rémi Cadene, Nicolas Thome, and Matthieu Cord.
\newblock Block: Bilinear superdiagonal fusion for visual question answering and visual relationship detection.
\newblock 2019.

\bibitem[Chen et~al.(2024)Chen, Ouyang, Gao, Chen, Chen, Wang, Zhang, Cai, Ji, Yu, Wan, and Wang]{huatuogptvision}
Junying Chen, Ruyi Ouyang, Anningzhe Gao, Shunian Chen, Guiming~Hardy Chen, Xidong Wang, Ruifei Zhang, Zhenyang Cai, Ke Ji, Guangjun Yu, Xiang Wan, and Benyou Wang.
\newblock Huatuogpt-vision, towards injecting medical visual knowledge into multimodal llms at scale, 2024.

\bibitem[Chen et~al.(2022)Chen, Du, Hu, Liu, Li, Wan, and Chang]{m3ae}
Zhihong Chen, Yuhao Du, Jinpeng Hu, Yang Liu, Guanbin Li, Xiang Wan, and Tsung-Hui Chang.
\newblock Multi-modal masked autoencoders for medical vision-and-language pre-training.
\newblock In \emph{International Conference on Medical Image Computing and Computer-Assisted Intervention}. Springer, 2022.

\bibitem[Cho et~al.(2023)Cho, Kim, Ryu, and Kweon]{vqa_object1}
Jae~Won Cho, Dong-Jin Kim, Hyeonggon Ryu, and In~So Kweon.
\newblock Generative bias for robust visual question answering.
\newblock In \emph{Proceedings of the IEEE/CVF Conference on Computer Vision and Pattern Recognition (CVPR)}, pages 11681--11690, 2023.

\bibitem[Cong et~al.(2022)Cong, Xu, Guo, and Tian]{mtl}
Fuze Cong, Shibiao Xu, Li Guo, and Yinbing Tian.
\newblock Caption-aware medical vqa via semantic focusing and progressive cross-modality comprehension.
\newblock In \emph{Proceedings of the 30th ACM International Conference on Multimedia}, page 3569–3577, New York, NY, USA, 2022. Association for Computing Machinery.

\bibitem[Devlin et~al.(2019)Devlin, Chang, Lee, and Toutanova]{bert}
Jacob Devlin, Ming-Wei Chang, Kenton Lee, and Kristina Toutanova.
\newblock {BERT}: Pre-training of deep bidirectional transformers for language understanding.
\newblock In \emph{Proceedings of the 2019 Conference of the North {A}merican Chapter of the Association for Computational Linguistics: Human Language Technologies, Volume 1 (Long and Short Papers)}, pages 4171--4186, Minneapolis, Minnesota, 2019. Association for Computational Linguistics.

\bibitem[Eslami et~al.(2023)Eslami, Meinel, and de~Melo]{pubmedclip}
Sedigheh Eslami, Christoph Meinel, and Gerard de Melo.
\newblock {P}ub{M}ed{CLIP}: How much does {CLIP} benefit visual question answering in the medical domain?
\newblock In \emph{Findings of the Association for Computational Linguistics: EACL 2023}, pages 1181--1193, Dubrovnik, Croatia, 2023. Association for Computational Linguistics.

\bibitem[Fang et~al.(2023)Fang, Jose, Jain, Schmidt, Toshev, and Shankar]{dfn_vit}
Alex Fang, Albin~Madappally Jose, Amit Jain, Ludwig Schmidt, Alexander Toshev, and Vaishaal Shankar.
\newblock Data filtering networks, 2023.

\bibitem[He et~al.(2020)He, Zhang, Mou, Xing, and Xie]{pathvqa}
Xuehai He, Yichen Zhang, Luntian Mou, Eric Xing, and Pengtao Xie.
\newblock Pathvqa: 30000+ questions for medical visual question answering, 2020.

\bibitem[Hu et~al.(2022)Hu, yelong shen, Wallis, Allen-Zhu, Li, Wang, Wang, and Chen]{lora}
Edward~J Hu, yelong shen, Phillip Wallis, Zeyuan Allen-Zhu, Yuanzhi Li, Shean Wang, Lu Wang, and Weizhu Chen.
\newblock Lo{RA}: Low-rank adaptation of large language models.
\newblock In \emph{International Conference on Learning Representations}, 2022.

\bibitem[Jiang et~al.(2023)Jiang, Sablayrolles, Mensch, Bamford, Chaplot, de~las Casas, Bressand, Lengyel, Lample, Saulnier, Lavaud, Lachaux, Stock, Scao, Lavril, Wang, Lacroix, and Sayed]{mistral}
Albert~Q. Jiang, Alexandre Sablayrolles, Arthur Mensch, Chris Bamford, Devendra~Singh Chaplot, Diego de~las Casas, Florian Bressand, Gianna Lengyel, Guillaume Lample, Lucile Saulnier, Lélio~Renard Lavaud, Marie-Anne Lachaux, Pierre Stock, Teven~Le Scao, Thibaut Lavril, Thomas Wang, Timothée Lacroix, and William~El Sayed.
\newblock Mistral 7b, 2023.

\bibitem[Krishna et~al.(2017)Krishna, Zhu, Groth, Johnson, Hata, Kravitz, Chen, Kalantidis, Li, Shamma, Bernstein, and Fei-Fei]{visual_genome}
Ranjay Krishna, Yuke Zhu, Oliver Groth, Justin Johnson, Kenji Hata, Joshua Kravitz, Stephanie Chen, Yannis Kalantidis, Li-Jia Li, David~A. Shamma, Michael~S. Bernstein, and Li Fei-Fei.
\newblock Visual genome: Connecting language and vision using crowdsourced dense image annotations.
\newblock \emph{Int. J. Comput. Vision}, 123\penalty0 (1):\penalty0 32–73, 2017.

\bibitem[Lau et~al.(2018)Lau, Gayen, Ben~Abacha, and Demner-Fushman]{vqa-rad}
Jason~J Lau, Soumya Gayen, Asma Ben~Abacha, and Dina Demner-Fushman.
\newblock A dataset of clinically generated visual questions and answers about radiology images.
\newblock \emph{Scientific data}, 5\penalty0 (1):\penalty0 1--10, 2018.

\bibitem[Li et~al.(2023{\natexlab{a}})Li, Wong, Zhang, Usuyama, Liu, Yang, Naumann, Poon, and Gao]{llavamed}
Chunyuan Li, Cliff Wong, Sheng Zhang, Naoto Usuyama, Haotian Liu, Jianwei Yang, Tristan Naumann, Hoifung Poon, and Jianfeng Gao.
\newblock {LL}a{VA}-med: Training a large language-and-vision assistant for biomedicine in one day.
\newblock In \emph{Thirty-seventh Conference on Neural Information Processing Systems Datasets and Benchmarks Track}, 2023{\natexlab{a}}.

\bibitem[Li et~al.(2023{\natexlab{b}})Li, Li, Savarese, and Hoi]{blip2}
Junnan Li, Dongxu Li, Silvio Savarese, and Steven Hoi.
\newblock Blip-2: Bootstrapping language-image pre-training with frozen image encoders and large language models, 2023{\natexlab{b}}.

\bibitem[Li et~al.(2024)Li, Skinner, Yang, Quaranto, Schwaitzberg, Kim, and Xiong]{llava_surg}
Jiajie Li, Garrett Skinner, Gene Yang, Brian~R Quaranto, Steven~D Schwaitzberg, Peter C~W Kim, and Jinjun Xiong.
\newblock Llava-surg: Towards multimodal surgical assistant via structured surgical video learning, 2024.

\bibitem[Li et~al.(2019)Li, Yatskar, Yin, Hsieh, and Chang]{visualbert}
Liunian~Harold Li, Mark Yatskar, Da Yin, Cho-Jui Hsieh, and Kai-Wei Chang.
\newblock Visualbert: A simple and performant baseline for vision and language, 2019.

\bibitem[Li et~al.(2023{\natexlab{c}})Li, Liu, He, Zhao, and Zhong]{MUMC}
Pengfei Li, Gang Liu, Jinlong He, Zixu Zhao, and Shenjun Zhong.
\newblock Masked vision and language pre-training with unimodal and multimodal contrastive losses for medical visual question answering.
\newblock pages 374--383, 2023{\natexlab{c}}.

\bibitem[Lin et~al.(2024)Lin, Hu, Fu, Yang, Chng, Kawasaki, Chui, and Liu]{detection}
Wenjun Lin, Yan Hu, Huazhu Fu, Mingming Yang, Chin-Boon Chng, Ryo Kawasaki, Cheekong Chui, and Jiang Liu.
\newblock Instrument-tissue interaction detection framework for surgical video understanding.
\newblock \emph{IEEE Transactions on Medical Imaging}, pages 1--1, 2024.

\bibitem[Liu et~al.(2021)Liu, Zhan, Xu, Ma, Yang, and Wu]{slake}
Bo Liu, Li-Ming Zhan, Li Xu, Lin Ma, Yan Yang, and Xiao-Ming Wu.
\newblock Slake: A semantically-labeled knowledge-enhanced dataset for medical visual question answering.
\newblock In \emph{2021 IEEE 18th International Symposium on Biomedical Imaging (ISBI)}, pages 1650--1654, 2021.

\bibitem[Liu et~al.(2024)Liu, Tian, Chen, Song, and Zhang]{r2llm}
Chang Liu, Yuanhe Tian, Weidong Chen, Yan Song, and Yongdong Zhang.
\newblock Bootstrapping large language models for radiology report generation.
\newblock \emph{Proceedings of the AAAI Conference on Artificial Intelligence}, 38\penalty0 (17):\penalty0 18635--18643, 2024.

\bibitem[Liu et~al.(2023)Liu, Li, Wu, and Lee]{llava}
Haotian Liu, Chunyuan Li, Qingyang Wu, and Yong~Jae Lee.
\newblock Visual instruction tuning.
\newblock In \emph{Thirty-seventh Conference on Neural Information Processing Systems}, 2023.

\bibitem[Moor et~al.(2023)Moor, Huang, Wu, Yasunaga, Zakka, Dalmia, Reis, Rajpurkar, and Leskovec]{medflamingo}
Michael Moor, Qian Huang, Shirley Wu, Michihiro Yasunaga, Cyril Zakka, Yash Dalmia, Eduardo~Pontes Reis, Pranav Rajpurkar, and Jure Leskovec.
\newblock Med-flamingo: a multimodal medical few-shot learner, 2023.

\bibitem[Nguyen et~al.(2019)Nguyen, Do, Nguyen, Do, Tjiputra, and Tran]{aioz_mevf_miccai19}
Binh~D. Nguyen, Thanh-Toan Do, Binh~X. Nguyen, Tuong Do, Erman Tjiputra, and Quang~D. Tran.
\newblock Overcoming data limitation in medical visual question answering.
\newblock In \emph{MICCAI}, 2019.

\bibitem[Nwoye et~al.(2022)Nwoye, Yu, Gonzalez, Seeliger, Mascagni, Mutter, Marescaux, and Padoy]{cholect45}
Chinedu~Innocent Nwoye, Tong Yu, Cristians Gonzalez, Barbara Seeliger, Pietro Mascagni, Didier Mutter, Jacques Marescaux, and Nicolas Padoy.
\newblock Rendezvous: Attention mechanisms for the recognition of surgical action triplets in endoscopic videos.
\newblock \emph{Medical Image Analysis}, 78:\penalty0 102433, 2022.

\bibitem[OpenAI(2023)]{2023GPT4VisionSC}
OpenAI.
\newblock Gpt-4v(ision) system card.
\newblock 2023.

\bibitem[OpenAI(2024)]{gpt4o}
OpenAI.
\newblock Hello gpt-4o.
\newblock 2024.

\bibitem[Radford et~al.(2019)Radford, Wu, Child, Luan, Amodei, Sutskever, et~al.]{gpt2}
Alec Radford, Jeffrey Wu, Rewon Child, David Luan, Dario Amodei, Ilya Sutskever, et~al.
\newblock Language models are unsupervised multitask learners.
\newblock \emph{OpenAI blog}, 1\penalty0 (8):\penalty0 9, 2019.

\bibitem[Radford et~al.(2021)Radford, Kim, Hallacy, Ramesh, Goh, Agarwal, Sastry, Askell, Mishkin, Clark, Krueger, and Sutskever]{clip}
Alec Radford, Jong~Wook Kim, Chris Hallacy, Aditya Ramesh, Gabriel Goh, Sandhini Agarwal, Girish Sastry, Amanda Askell, Pamela Mishkin, Jack Clark, Gretchen Krueger, and Ilya Sutskever.
\newblock Learning transferable visual models from natural language supervision, 2021.

\bibitem[Seenivasan et~al.(2022)Seenivasan, Islam, Krishna, and Ren]{sugvqa}
Lalithkumar Seenivasan, Mobarakol Islam, Adithya~K. Krishna, and Hongliang Ren.
\newblock Surgical-vqa: Visual question answering in surgical scenes using transformer.
\newblock In \emph{Medical Image Computing and Computer Assisted Intervention -- MICCAI 2022}, pages 33--43, Cham, 2022. Springer Nature Switzerland.

\bibitem[Seenivasan et~al.(2023)Seenivasan, Islam, Kannan, and Ren]{suggpt}
Lalithkumar Seenivasan, Mobarakol Islam, Gokul Kannan, and Hongliang Ren.
\newblock Surgicalgpt: End-to-end language-vision gpt for visual question answering in surgery.
\newblock In \emph{Medical Image Computing and Computer Assisted Intervention -- MICCAI 2023}, pages 281--290, Cham, 2023. Springer Nature Switzerland.

\bibitem[Sparck~Jones(1988)]{tfidf}
Karen Sparck~Jones.
\newblock \emph{A statistical interpretation of term specificity and its application in retrieval}, page 132–142.
\newblock Taylor Graham Publishing, GBR, 1988.

\bibitem[Touvron et~al.(2023)Touvron, Lavril, Izacard, Martinet, Lachaux, Lacroix, Rozière, Goyal, Hambro, Azhar, Rodriguez, Joulin, Grave, and Lample]{llama}
Hugo Touvron, Thibaut Lavril, Gautier Izacard, Xavier Martinet, Marie-Anne Lachaux, Timothée Lacroix, Baptiste Rozière, Naman Goyal, Eric Hambro, Faisal Azhar, Aurelien Rodriguez, Armand Joulin, Edouard Grave, and Guillaume Lample.
\newblock Llama: Open and efficient foundation language models, 2023.

\bibitem[Twinanda et~al.(2017)Twinanda, Shehata, Mutter, Marescaux, de~Mathelin, and Padoy]{cholec80}
Andru~P. Twinanda, Sherif Shehata, Didier Mutter, Jacques Marescaux, Michel de Mathelin, and Nicolas Padoy.
\newblock Endonet: A deep architecture for recognition tasks on laparoscopic videos.
\newblock \emph{IEEE Transactions on Medical Imaging}, 36\penalty0 (1):\penalty0 86--97, 2017.

\bibitem[Valderrama et~al.(2022)Valderrama, Ruiz, Hern{\'a}ndez, Ayobi, Verlyck, Santander, Caicedo, Fern{\'a}ndez, and Arbel{\'a}ez]{psi-ava}
Natalia Valderrama, Paola Ruiz, Isabela Hern{\'a}ndez, Nicol{\'a}s Ayobi, Mathilde Verlyck, Jessica Santander, Juan Caicedo, Nicol{\'a}s Fern{\'a}ndez, and Pablo Arbel{\'a}ez.
\newblock Towards holistic surgical scene understanding.
\newblock In \emph{Medical Image Computing and Computer Assisted Intervention -- MICCAI 2022}, pages 442--452, Cham, 2022. Springer Nature Switzerland.

\bibitem[Xia et~al.(2024)Xia, Zhu, Li, Zhu, Li, Li, Zhang, and Yao]{rule}
Peng Xia, Kangyu Zhu, Haoran Li, Hongtu Zhu, Yun Li, Gang Li, Linjun Zhang, and Huaxiu Yao.
\newblock Rule: Reliable multimodal rag for factuality in medical vision language models.
\newblock \emph{arXiv preprint arXiv:2407.05131}, 2024.

\bibitem[Xue et~al.(2024)Xue, Shu, Awadalla, Wang, Yan, Purushwalkam, Zhou, Prabhu, Dai, Ryoo, Kendre, Zhang, Qin, Zhang, Chen, Yu, Tan, Awalgaonkar, Heinecke, Wang, Choi, Schmidt, Chen, Savarese, Niebles, Xiong, and Xu]{blip3}
Le Xue, Manli Shu, Anas Awadalla, Jun Wang, An Yan, Senthil Purushwalkam, Honglu Zhou, Viraj Prabhu, Yutong Dai, Michael~S Ryoo, Shrikant Kendre, Jieyu Zhang, Can Qin, Shu Zhang, Chia-Chih Chen, Ning Yu, Juntao Tan, Tulika~Manoj Awalgaonkar, Shelby Heinecke, Huan Wang, Yejin Choi, Ludwig Schmidt, Zeyuan Chen, Silvio Savarese, Juan~Carlos Niebles, Caiming Xiong, and Ran Xu.
\newblock xgen-mm (blip-3): A family of open large multimodal models, 2024.

\bibitem[Yang et~al.(2024)Yang, Yang, Hui, Zheng, Yu, Zhou, Li, Li, Liu, Huang, Dong, Wei, Lin, Tang, Wang, Yang, Tu, Zhang, Ma, Yang, Xu, Zhou, Bai, He, Lin, Dang, Lu, Chen, Yang, Li, Xue, Ni, Zhang, Wang, Peng, Men, Gao, Lin, Wang, Bai, Tan, Zhu, Li, Liu, Ge, Deng, Zhou, Ren, Zhang, Wei, Ren, Liu, Fan, Yao, Zhang, Wan, Chu, Liu, Cui, Zhang, Guo, and Fan]{qwen2}
An Yang, Baosong Yang, Binyuan Hui, Bo Zheng, Bowen Yu, Chang Zhou, Chengpeng Li, Chengyuan Li, Dayiheng Liu, Fei Huang, Guanting Dong, Haoran Wei, Huan Lin, Jialong Tang, Jialin Wang, Jian Yang, Jianhong Tu, Jianwei Zhang, Jianxin Ma, Jianxin Yang, Jin Xu, Jingren Zhou, Jinze Bai, Jinzheng He, Junyang Lin, Kai Dang, Keming Lu, Keqin Chen, Kexin Yang, Mei Li, Mingfeng Xue, Na Ni, Pei Zhang, Peng Wang, Ru Peng, Rui Men, Ruize Gao, Runji Lin, Shijie Wang, Shuai Bai, Sinan Tan, Tianhang Zhu, Tianhao Li, Tianyu Liu, Wenbin Ge, Xiaodong Deng, Xiaohuan Zhou, Xingzhang Ren, Xinyu Zhang, Xipin Wei, Xuancheng Ren, Xuejing Liu, Yang Fan, Yang Yao, Yichang Zhang, Yu Wan, Yunfei Chu, Yuqiong Liu, Zeyu Cui, Zhenru Zhang, Zhifang Guo, and Zhihao Fan.
\newblock Qwen2 technical report, 2024.

\bibitem[Yang et~al.(2016)Yang, He, Gao, Deng, and Smola]{san}
Zichao Yang, Xiaodong He, Jianfeng Gao, Li Deng, and Alex Smola.
\newblock Stacked attention networks for image question answering.
\newblock In \emph{Proceedings of the IEEE Conference on Computer Vision and Pattern Recognition (CVPR)}, 2016.

\bibitem[Yu et~al.(2017)Yu, Yu, Fan, and Tao]{mfb}
Zhou Yu, Jun Yu, Jianping Fan, and Dacheng Tao.
\newblock Multi-modal factorized bilinear pooling with co-attention learning for visual question answering.
\newblock In \emph{2017 IEEE International Conference on Computer Vision (ICCV)}, pages 1839--1848, 2017.

\bibitem[Yu et~al.(2018)Yu, Yu, Xiang, Fan, and Tao]{mfh}
Zhou Yu, Jun Yu, Chenchao Xiang, Jianping Fan, and Dacheng Tao.
\newblock Beyond bilinear: Generalized multimodal factorized high-order pooling for visual question answering.
\newblock \emph{IEEE TNNLS}, 29\penalty0 (12):\penalty0 5947--5959, 2018.

\bibitem[Yuan et~al.(2024)Yuan, Kattel, Lavanchy, Navab, Srivastav, and Padoy]{ssgvqa}
Kun Yuan, Manasi Kattel, Joel~L. Lavanchy, Nassir Navab, Vinkle Srivastav, and Nicolas Padoy.
\newblock Advancing surgical vqa with scene graph knowledge, 2024.

\bibitem[Zhan et~al.(2020)Zhan, Liu, Fan, Chen, and Wu]{cond_reason}
Li-Ming Zhan, Bo Liu, Lu Fan, Jiaxin Chen, and Xiao-Ming Wu.
\newblock Medical visual question answering via conditional reasoning.
\newblock In \emph{Proceedings of the 28th ACM International Conference on Multimedia}, page 2345–2354, New York, NY, USA, 2020. Association for Computing Machinery.

\bibitem[Zhang et~al.(2024)Zhang, Wu, Zhao, Lin, Zhang, Wang, and Xie]{pmc_vqa}
Xiaoman Zhang, Chaoyi Wu, Ziheng Zhao, Weixiong Lin, Ya Zhang, Yanfeng Wang, and Weidi Xie.
\newblock Pmc-vqa: Visual instruction tuning for medical visual question answering, 2024.

\bibitem[Zhu et~al.(2024)Zhu, Chen, Shen, Li, and Elhoseiny]{minigpt}
Deyao Zhu, Jun Chen, Xiaoqian Shen, Xiang Li, and Mohamed Elhoseiny.
\newblock Mini{GPT}-4: Enhancing vision-language understanding with advanced large language models.
\newblock In \emph{The Twelfth International Conference on Learning Representations}, 2024.

\end{thebibliography}
